%% file: main.tex
\ifdefined\pdfminorversion
\fi

\documentclass[11pt,a4paper,logo]{lumia}
\usepackage[numbers,sort&compress]{natbib}
\input{preamble}

\definecolor{infrecred}{HTML}{D2071E}
\definecolor{abstrabg}{HTML}{FCDDDC}
\colorlet{abstradark}{infrecred}
\definecolor{reportaccent}{HTML}{BB0819}
\colorlet{reporttoc}{reportaccent}
\definecolor{cayenne}{HTML}{941100}
\hypersetup{allcolors=cayenne,pdftitle={AAAWorld: How Far Are We From Generating AAA Games With a Single Prompt?}}
\setheadertext{\pubedit{AAAWorld}}
\setheadertitle{\pubedit{AAAWorld}}
\title{{\fontsize{16}{20}\selectfont \pubedit{AAAWorld: How Far Are We From Generating AAA Games With a Single Prompt?}\par}}
\author{\shortstack[c]{{\fontsize{13pt}{15pt}\selectfont AAAWorld Team}\\
  {\normalfont\small\href{https://bench.cardinal-agi.com/aaaworld}{\pubedit{Project Page: \enspace\nolinkurl{bench.cardinal-agi.com/aaaworld}}}}}}

\begin{document}
\input{sec/0_abstract}
\teaserfigure{\input{fig/teaser}}
\maketitle

\clearpage
\begingroup
\hypersetup{linkcolor=reporttoc,linktoc=section}
\setlength{\parskip}{0pt}
\setcounter{tocdepth}{2}
\tableofcontents
\endgroup
\clearpage

\input{sec/1_intro}

\input{sec/2_related}
\input{sec/3_benchmark}
\input{sec/6_results}
\input{sec/8_conclusion}
\input{sec/9_llm_usage}

\clearpage
\input{sec/8_authors}

\clearpage
\phantomsection
\addcontentsline{toc}{section}{References}
\bibliography{main}

\clearpage
\appendix
\input{sec/X_suppl}
\end{document}

%% file: preamble.tex
\usepackage{xspace,subcaption}
\usepackage[capitalize,noabbrev]{cleveref}

\usepackage{currfile}
\usepackage{multirow}
\usepackage{tabularx}
\usepackage{placeins}
\usepackage{colortbl}
\usepackage{pifont}
\usepackage[normalem]{ulem}
\usepackage{pgfplots}
\pgfplotsset{compat=1.18}

\DeclareRobustCommand{\pubedit}[1]{#1}
\DeclareCaptionLabelFormat{pubappendix}{\pubedit{#1~#2}}
\DeclareRobustCommand{\pubapp}[1]{\hyperref[#1]{\pubedit{Appendix~\ref*{#1}}}}
\newcommand{\ours}{\pubedit{AAAWorld}\xspace}
\newcommand{\Ours}{\ours}

\newcommand{\tabhead}[1]{%
  \begin{tabular}[b]{@{}c@{}}
    #1
  \end{tabular}%
}

\definecolor{rqbackground}{HTML}{F4F6F8}
\definecolor{rqaccent}{HTML}{2C6490}

\definecolor{tableyescolor}{HTML}{187435}
\definecolor{tablepartialcolor}{HTML}{C4770B}
\definecolor{tablenocolor}{HTML}{BD3434}

\DeclareRobustCommand{\tableyes}{%
  \textcolor{tableyescolor}{\ding{51}}%
}

\DeclareRobustCommand{\tablepartial}{%
  \textcolor{tablepartialcolor}{\ensuremath{\bigcirc}}%
}

\DeclareRobustCommand{\tableno}{%
  \textcolor{tablenocolor}{\ding{55}}%
}

\newtcolorbox{researchquestion}{
  sharp corners, colback=rqbackground, colframe=rqaccent,
  boxrule=0pt, leftrule=1.2pt, boxsep=0pt,
  left=8pt, right=8pt, top=6pt, bottom=6pt,
  before skip=6pt, after skip=6pt,
  fontupper=\itshape,
  before upper={%
    \settowidth{\leftskip}{\textup{RQ1.} }%
    \setlength{\parindent}{-\leftskip}%
    \setlength{\parskip}{2pt}%
    \hyphenpenalty=10000\relax
    \indent
  }
}

\usepackage{CJKutf8}

\usepackage{microtype}

%% file: sec/0_abstract.tex
\begin{abstract}
As Large Language Models (LLMs) advance in programming, complex game generation has emerged as a challenging testbed for \pubedit{spatial reasoning capability}. However, evaluating generated games beyond runnability remains difficult: existing benchmarks rely mainly on programmatic checks or LLM-as-judges, overlooking the open-ended, subjective nature of player experience. We introduce \textbf{\pubedit{AAAWorld}}, a human-grounded benchmark of 1,001 games across 15 primary and 42 secondary genres. Our play-before-judge protocol pairs synchronized gameplay recordings and inputs with human ratings to measure \textbf{Reference Attainment (RA)}, the percentage of quality achieved relative to the reference game. We propose an \pubedit{Efficient Elimination Evaluation (EEE) mechanism to facilitate game generation evaluation} across Easy, Medium, and Hard tasks, plus a 15-task challenge comparing five harnesses and four-round feedback strategies. In the main study, certified annotators contributed 11,794 qualified evaluations over 979 playable versions across eight models and five harnesses. GPT-6 Astra achieves the highest RA in all three OpenCode tiers and outperforms Opus~5.5 on Easy and Medium despite lower Auto checklist scores. Multi-round analysis shows player feedback outperforms \pubedit{Reward-driven Self-Improvement (RSI) guided by Auto reports} across both playability bottlenecks and overall experience. \pubedit{Our trained Game Reviewer Agent (GRA) produces scores closer to human judgments than those of the frozen baselines, with} DEV/TEST mean absolute errors of 6.92/7.45 when scoring shared human replays (GRA-35B) and 7.58/7.85 when playing and scoring its own trajectories (GRA-GRPO-9B).
\end{abstract}

%% file: fig/teaser.tex
\noindent
\begin{minipage}{1.0\textwidth}
    \captionsetup{type=figure,font=small}
    \centering
    \includegraphics[width=0.999\linewidth,keepaspectratio]{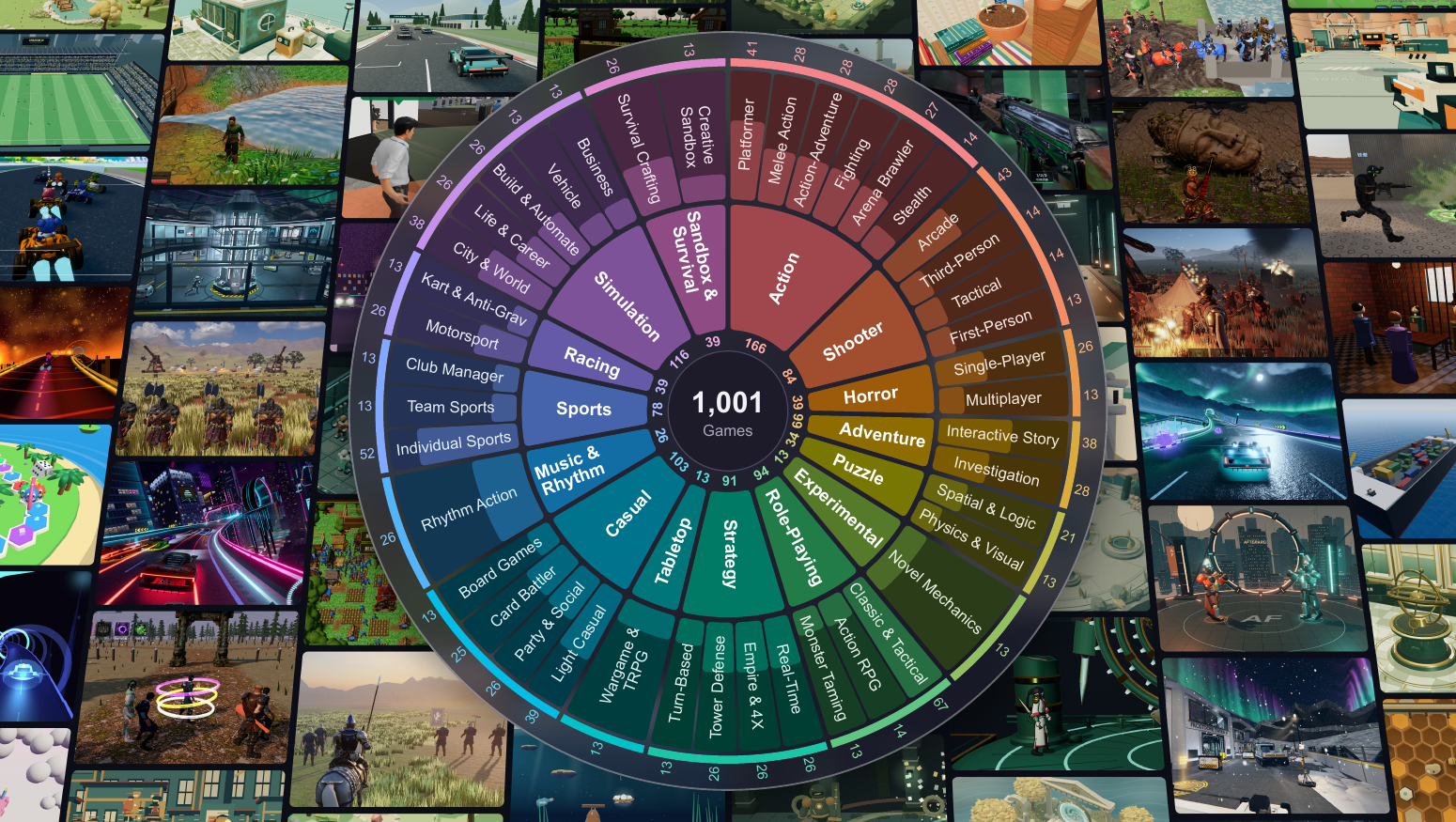}
    \caption{Game taxonomy and representative \textbf{AI-generated game examples} extracted from \pubedit{AAAWorld}. The benchmark comprises 1,001 Steam-referenced games across 15 primary and 42 secondary genres.}
    \label{fig:intro}
\end{minipage}
\par\vspace{6pt}

%% file: sec/1_intro.tex
\begingroup
\clubpenalty=10000
\widowpenalty=10000
\section{Introduction}
\label{sec:intro}

As LLMs have made remarkable progress in automated coding, synthesizing complex games from natural-language prompts has become a demanding testbed for \pubedit{spatial reasoning capability}~\cite{lu2026worldcoder,jiang2026opengame,zhang2026webgamebench}. Unlike conventional code tasks, game generation requires synthesizing code logic, aesthetics, gameplay mechanics, and control feedback into a coherent, engaging experience. Simple execution-based runnability checks are therefore insufficient. Two functionally runnable games may differ substantially in visuals, mechanical depth, and control feel. Measuring genuine progress needs to move beyond naive programmatic verification.


However, evaluating generated games beyond runnability remains a challenge. Game generation is inherently open-ended and lacks a single ground-truth answer. Existing benchmarks largely rely on programmatic checks or LLM-as-judges to scale their evaluations. Programmatic checks verify functional rather than perceived game quality~\cite{chi2026gamedevbench}, and zero-shot multimodal or agentic judges can diverge from human preferences~\cite{zhuge2025agentjudge,chen2024mllmjudge,panickssery2024llm} or suffer from weak autonomous play~\cite{jia2026gamegenverifier}. Grounding evaluation in real human playtesting is essential, yet collecting reliable human judgments at scale requires careful protocol design and resource allocation.

To address these challenges, we introduce \textbf{\pubedit{AAAWorld}}, a human-grounded benchmark anchored in 1,001 Steam-referenced games across 15 primary and 42 secondary genres. Under our \emph{play-before-judge} evaluation protocol, screened playtesters inspect the reference context and play each generated game, with screen recordings and input trajectories recorded alongside their ratings. We define \emph{Reference Attainment} (RA) as the percentage of human-judged quality relative to the reference game, complemented by S-D grades for Stability, Playability, Visual Quality, and Interactive Feedback.

After generating initial builds across all 1,001 games with Gemini~3.7 Flash and scoring them via Auto, we randomly sample 120 tasks across the three evaluation tiers plus 15 challenge tasks to form the 135-task open set for our main studies, while reserving the remaining pool as a dynamically rotated closed set. To focus human playtesting on stronger frontier generators, \pubedit{Efficient Elimination Evaluation (EEE)} is proposed for narrowing OpenCode comparisons across the Easy, Medium, and Hard sets, while a separate 15-task challenge compares five harnesses and four-round feedback strategies (Auto reports, developer critiques, and player feedback). The main study covers 8 models, 5 harnesses, 979 playable versions, and 11,794 qualified ratings. We ask three research questions:

\begin{researchquestion}
{\upshape RQ1.} How do coding models and agent harnesses compare in experiential quality relative to reference games?\par
{\upshape RQ2.} How do feedback strategies affect human-rated quality over iterative development rounds?\par
{\upshape RQ3.} How well do automated graders align with human judges, and can the reviewer agent generalize to unseen tasks?
\end{researchquestion}

Regarding \textbf{RQ1}, within-tier OpenCode comparisons reveal significant quality gaps: GPT-6 Astra has the highest RA in all three tiers and outperforms Opus~5.5 on Easy and Medium despite lower Auto scores, whereas Opus~5.5 leads on the separate 15-task multi-harness challenge. This confirms functional checklists are an insufficient proxy for quality.

Regarding \textbf{RQ2}, four-round iteration shows that human-in-the-loop player feedback outperforms \pubedit{objective-feedback RSI (Obj-RSI)} at R4 by up to 8.92 RA points, whereas \pubedit{Obj-RSI} lowers human-rated RA. Experiential dimension analysis shows human feedback lifts both initial bottlenecks and non-bottleneck dimensions.

Regarding \textbf{RQ3}, programmatic Auto scores correlate weakly with human consensus across the 979 versions ($\rho=0.174$). Our \pubedit{trained} \textbf{Game Reviewer Agent (GRA)} orders human-rated version pairs with statistically separated scores at 83.26\%/77.42\% accuracy on DEV/TEST when scoring shared replays (GRA-35B), and at 71.67\%/69.75\% when playing and scoring its own trajectories (GRA-GRPO-9B).

\begin{table*}[t]
\centering
\caption[Comparison of evaluation features.]{\textbf{Comparison of evaluation features in game and interactive code generation.}
\tableyes{} indicates included; \tableno{} indicates not included or not reported.
Human: manual scoring (\tableyes), other human involvement (\tablepartial);
Trained: human-supervised judge training protocol; Iter.: multi-round iteration analysis;
Feedback: comparison of different feedback strategies.}
\label{tab:related}
\footnotesize
\setlength{\tabcolsep}{3pt}
\renewcommand{\arraystretch}{1.18}
\begin{tabularx}{\linewidth}{@{}lcc*{4}{>{\centering\arraybackslash}X}@{}}
\toprule
\textbf{Work} & \textbf{Engine} & \textbf{Content} & \textbf{Human} & \textbf{Trained} & \textbf{Iter.} & \textbf{Feedback} \\
\midrule
ArtifactsBench~\cite{zhang2025artifactsbench} & Not specified & 2D/3D & \tablepartial & \tableno & \tableno & \tableno \\
V-GameGym~\cite{zhang2026vgamegym} & Pygame & 2D & \tablepartial & \tableno & \tableno & \tableno \\
WorldCoder-Bench~\cite{lu2026worldcoder} & Three.js & 3D & \tablepartial & \tableno & \tableno & \tableno \\
WebGameBench~\cite{zhang2026webgamebench} & Not specified & 2D & \tablepartial & \tableno & \tableno & \tableno \\
GameCraft-Bench~\cite{luo2026gamecraft} & Godot & 2D & \tablepartial & \tableno & \tableno & \tableno \\
SWE-Game~\cite{chen2026swegame} & Godot/Unity & 2D/3D & \tablepartial & \tableno & \tableno & \tableno \\
GameGen-Verifier~\cite{jia2026gamegenverifier} & Not specified & 2D/3D & \tablepartial & \tableno & \tableno & \tableno \\
\addlinespace[2pt]
OpenGame~\cite{jiang2026opengame} & Phaser & 2D & \tablepartial & \tableno & \tableno & \tableyes \\
GameXpert-Bench~\cite{chen2026gamexpert} & Unrestricted & 2D/3D & \tableyes & \tableno & \tableno & \tableyes \\
\addlinespace[2pt]
A2Z GameSpec-Bench~\cite{lee2026a2z} & Phaser/Three.js & 2D/3D & \tableno & \tableno & \tableyes & \tableyes \\
PlaytestArena~\cite{huang2026playtestarena} & Not specified & 2D & \tablepartial & \tableno & \tableyes & \tableyes \\
\midrule
\rowcolor{rqbackground}
\textbf{\Ours} & \textbf{Three.js} & \textbf{3D/2.5D} & \tableyes & \tableyes & \tableyes & \tableyes \\
\bottomrule
\end{tabularx}
\end{table*}

\medskip
\noindent Our main contributions are threefold:
\begin{enumerate}[topsep=3pt,itemsep=3pt]
    \item \textbf{A large-scale human-grounded benchmark.} We introduce \pubedit{AAAWorld} with a 1,001-game pool (a 135-task open set plus a dynamically rotated closed set), pairing our play-before-judge protocol and RA metric with multi-dimensional ratings.
    \item \textbf{Multi-round feedback and dimension analysis.} Across four iterative rounds, human-in-the-loop feedback outperforms \pubedit{Obj-RSI} at R4, lifting both initial bottlenecks and overall playability whereas \pubedit{Obj-RSI} degrades gameplay.
    \item \textbf{Human-calibrated automated evaluation.} \pubedit{We train GRA on gameplay trajectories and ratings. Its scores align more closely with human judgments than those of the frozen baselines across DEV and TEST, both on shared replays and in autonomous play.}
\end{enumerate}
\endgroup

%% file: sec/2_related.tex
\section{Related Work}
\label{sec:related}

\paragraph{Game generation.}
Foundation models increasingly span game simulation, code-based creation, and automated playtesting~\cite{luo2026ai}.
GameGAN~\cite{kim2020gamegan} and GameNGen~\cite{valevski2025gamengen} simulate games with adversarial and diffusion models; Genie~\cite{bruce2024genie} learns latent actions from videos without action labels, while WHAM~\cite{kanervisto2025wham} uses human gameplay and controller inputs.
Code-based approaches such as Holodeck~\cite{yang2024holodeck} and SceneCraft~\cite{hu2024scenecraft} generate scene layouts or Blender code, but do not produce complete games with rules, goals and feedback.
Recent coding-agent benchmarks cover web artifacts and Pygame games~\cite{zhang2025artifactsbench,zhang2026vgamegym}, browser games and Three.js worlds~\cite{jiang2026opengame,zhang2026webgamebench,lu2026worldcoder}, and Godot/Unity projects~\cite{luo2026gamecraft,chen2026swegame}.
Some compare model-harness combinations~\cite{luo2026gamecraft,zhang2026gameasg}, while feedback-driven systems guide iterative game revision~\cite{lee2026a2z,huang2026playtestarena,peng2026playcoder}.
\ours targets interactive 3D/2.5D scenes in Three.js, with quality anchored to commercial reference games within a specified scene scope.


\paragraph{Game evaluation.}
WorldCoder-Bench evaluates functional correctness~\cite{lu2026worldcoder}, while SWE-Game and GameGen-Verifier use human judgments to validate automated evaluation~\cite{chen2026swegame,jia2026gamegenverifier}.
In \ours, human play-before-judge ratings define the core quality metric, RA, and support reliability analysis.
Reference-anchored comparisons~\cite{si2025design2code}, human-trained visual evaluators~\cite{xu2023imagereward,he2024videoscore,li2025worldmodelbench}, and human--engine trajectory verification~\cite{zhou2026agentic} provide related foundations. We connect gameplay trajectories and ratings to GRA's action and assessment supervision, while validating functional checks against dense player judgments.
Feedback evaluation poses a further distinction: PlaytestArena's Play2Code and A2Z GameSpec-Bench rely on automated evaluators for both feedback and assessment~\cite{huang2026playtestarena,lee2026a2z}, whereas proxy optimization can diverge from human quality~\cite{gao2023scaling}. We instead evaluate feedback trajectories with multi-rater human playtesting of each revised build.

%% file: sec/3_benchmark.tex
\section{\ours}
\label{sec:benchmark}
\ours combines reference-game tasks, version-aligned human playtesting, and automated evaluation (\cref{fig:overview}). Full protocols are in \pubapp{sec:app-implementation}.
\input{fig/overview}

\subsection{Tasks and Evaluation Tracks}
\label{sec:task}\label{sec:pool}\label{sec:principles}\label{sec:tracks}
Each task prompt specifies the target subgenre and scoped core mechanics of a commercial reference game (released 2016--2025 across 15 primary and 42 secondary genres), asking an agent to build a standalone playable 3D/2.5D Three.js scene in its workspace with strictly original code, characters, UI, and assets; reference briefs and gameplay videos anchor playtester quality judgments. Generating initial builds across all 1,001 PC games with Gemini~3.7 Flash under OpenCode and scoring them via Auto (Codex with GPT-6 Astra), we randomly sample 120 tasks across the three evaluation tiers defined by initial-build Auto scores plus 15 challenge tasks to form our \textbf{135-task open set}, while retaining the remaining pool as a dynamically rotated \textbf{closed set}.

\paragraph{Generation and harness comparisons.}
\label{sec:screening}
Within the open set, \pubedit{EEE under OpenCode} covers 30 Easy, 40 Medium, and 50 Hard tasks across eight, four, and two models so playtesting focuses progressively on frontier generators. A separate 15-task challenge compares two models across five CLI/agent harnesses (OpenCode, Codex, Pi, Prime, and the DeepSeek CLI harness).

\paragraph{Feedback iteration.}
\label{sec:multiround}
On the 15 challenge tasks, Astra under Codex and OpenCode continues each prior workspace across R2--R4 under four prompt strategies: \emph{Fixed} issues staged level-addition, browser-playtest, and polish prompts; \emph{\pubedit{Obj-RSI}} supplies the prior Auto score, gate status, and item-level checklist report; \emph{HITL-Dev} provides concise critique notes on mechanics, modeling, and level design from experienced developers ($n\ge 3$); and \emph{HITL-Player} supplies aggregated player overall/four-dimension scores and playtest comments (templates in \pubapp{sec:app-prompt}). Sharing 30 R1 baselines yields 360 R2--R4 versions; the main study covers 135 tasks and 1,010 candidate versions.

\subsection{Human Playtesting and Post-Processing}
\label{sec:human}\label{sec:agg}
We recruited over 1,000 volunteers via crowdsourcing and multi-round qualification to retain 63 active annotators on the 979 playable versions in the main study (see \pubapp{sec:app-annotators}). Annotators first review each reference game's brief and video outside the session timer, then play the generated scene (recording 321.0 active in-game hours across 11,794 qualified sessions, or $\approx 1.6$\,min of active play per rating session, with a median of 73\,s and an IQR of 63--103\,s) and assign a Reference Attainment score $P\in[0,100]$ in five-point steps plus S--D grades for Stability, Playability, Visual Quality, and Interactive Feedback; mapping grades to $(25,50/3,25/3,25/6,0)$ and summing defines the Experience Composite $\mathrm{EC}\in[0,100]$.

After deduplicating to each rater's latest submission per version (12,136 raw sessions), hierarchical Student-$t$ residual outlier screening removes 342 flagged submissions and retains 11,794 qualified evaluations, providing at least ten ratings per version (harmonic replication $k_0=12.05$) with consensus reliability $\mathrm{ICC}(1,k_0)=0.727$~\cite{shrout1979intraclass}. Taking the unadjusted arithmetic mean $\bar P^H(g_{t,c})$ of retained raw ratings on each version $g_{t,c}$ (without subtracting rater offsets), configuration-level RA averages equally over tasks $t\in\mathcal T$:
\begin{equation}
\mathrm{RA}(c;\mathcal T)=\frac{1}{|\mathcal T|}\sum_{t\in\mathcal T}\bar P^H(g_{t,c}),
\label{eq:ra}
\end{equation}
assigning zero to 31 final gate failures in RA and $\mathrm{EC}$ while excluding them from grader-agreement analyses.

\subsection{Automated Evaluation and Validation}
\label{sec:eval}\label{sec:agreement}
\paragraph{Automated score (Auto).}
\label{sec:obj}
A Codex agent running GPT-6 Astra uses Playwright to test browser behavior. Six gates establish a reachable, nonblank interactive state without blocking errors; any failure yields zero. Passing artifacts are scored on 41 items in ten categories on a predefined functional checklist, normalized to 100 over applicable items.

\paragraph{Game Reviewer Agent (GRA).}
\label{sec:judge}
We split 255 reference-game families into 205 TRAIN, 25 DEV and 25 TEST families. Reviewers predict the rater-adjusted median human score for each version. On shared human replays, GRA-35B, frozen Qwen3.5-35B-A3B and frozen GPT-6 Astra each score the same 16 frames in one call. For autonomous play, we train GRA-GRPO-9B with group relative policy optimization. It and the frozen 9B and 35B Qwen baselines each play the game and score their own trajectory. We evaluated Astra on replays only due to the cost of autonomous play.

\paragraph{Agreement and statistical inference.}
\label{sec:transfer}
We evaluate grader alignment via instance-level correlations and within-task pairwise accuracy. Feedback gains are paired R4--R1 changes, regressions track adjacent-round drops of at least one median human tier in any dimension, and paired random-sign tests apply Holm correction within comparison families.

%% file: fig/overview.tex
\begin{figure*}[!t]
\centering
\includegraphics[width=\linewidth]{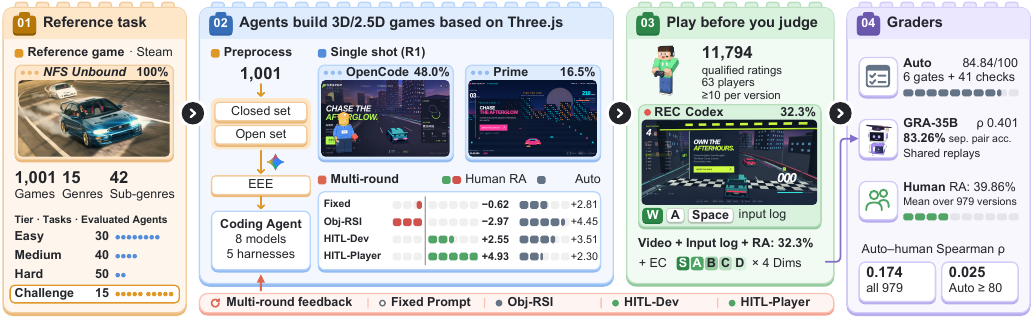}
\caption{\textbf{Overview of \ours}. Reference tasks lead to generated browser scenes, version-aligned human playtesting, and automated evaluation. After generating initial builds across the 1,001-game pool with Gemini~3.7 Flash and scoring them via Auto (Codex with GPT-6 Astra), we randomly sample 120 tasks across the three OpenCode tiers plus 15 challenge tasks to form the 135-task open set, while the dynamically rotated closed set reserves the remaining pool. Displayed screenshot scores are instance-level RA means; multi-round gains show Astra under OpenCode. The main study retained 11,794 qualified ratings across 979 playable versions. EC sums the four 0--25 human dimension scores and the GRA-35B panel reports DEV agreement on shared human replays.}
\label{fig:overview}
\end{figure*}

%% file: sec/6_results.tex
\input{tab/leaderboard}
\input{fig/harness_main}

\begin{figure*}[!t]
\centering
{\setlength{\fboxsep}{0.4pt}\setlength{\fboxrule}{0pt}%
\fbox{\includegraphics[width=\dimexpr.96\linewidth-2\fboxsep\relax]{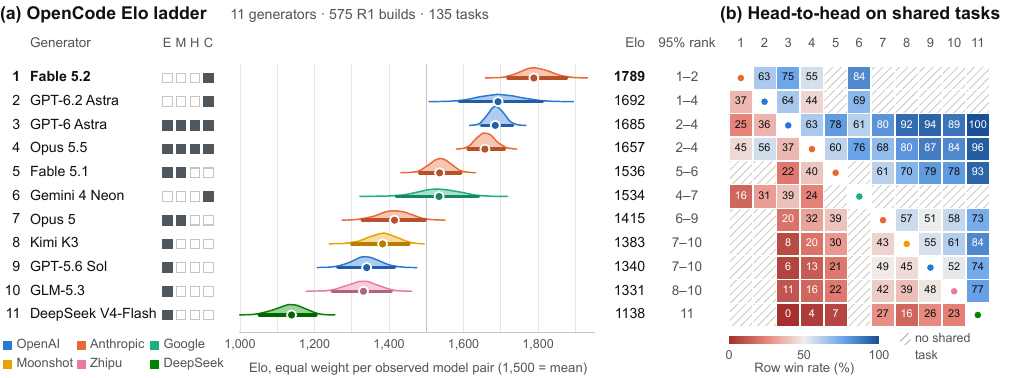}}}
\caption{\textbf{OpenCode Elo across 135 tasks.} (a) Bradley--Terry Elo fits RA win probabilities on shared tasks, weighting observed model pairs equally. Ridges and bars show 4,000 within-tier task bootstraps and 95\% intervals. Squares mark tier coverage. (b) Cells show the row model's estimated RA win rate on shared tasks. Hatched pairs never met.}
\label{fig:opencode-elo}
\end{figure*}

\section{Experiments and Analysis}
\label{sec:results}

\input{sec/5_setup}

\subsection{Single-Round Generation (RQ1)}
\label{sec:leaderboard}

Astra achieves the highest RA in each evaluation tier: 46.61\% on Easy, 48.21\% on Medium, and 42.71\% on Hard (\cref{tab:leaderboard}).
Its paired advantages over Opus~5.5 are 5.66 points on Easy (Holm-adjusted $p=0.0065$), 9.21 on Medium ($p=0.0003$), and 2.34 on Hard ($p=0.136$), establishing clear leads on Easy and Medium while Hard remains competitive.

Passing runtime checks does not imply comparable player experience.
All Easy artifacts from Astra and Opus~5.5 pass the gates, yet Opus~5.5 attains a higher Auto mean (86.75 versus 81.95) alongside substantially lower RA.
Across the eight Easy models, RA spans 13.66--46.61\%, exposing wide experiential gaps beyond executability.
On Easy, Astra leads Opus~5.5 in Stability (10.57 vs.\ 8.61), Visual Quality (8.43 vs.\ 6.54), and Interactive Feedback (8.85 vs.\ 7.11); on Hard, Astra retains a visual edge (7.98 vs.\ 6.69) while Opus~5.5 is slightly higher in Playability (8.11 vs.\ 7.91).

\paragraph{Role of the Harness.}
\label{sec:harness}
{\normalcolor With OpenCode fixed on 15 Challenge tasks, the three additional generators attain 48.07\% RA (Fable~5.2), 45.01\% (GPT-6.2 Astra) and 38.18\% (Gemini~4 Neon) (\cref{tab:harness}a). Across five harnesses, Opus~5.5 leads GPT-6 Astra (40.52--46.71\% vs.\ 31.95--41.85\% RA). Both peak under OpenCode (\cref{tab:harness}b). Only GPT-6 Astra has gate failures in this two-model comparison (one under Pi, two under Prime).}

{\normalcolor Fable~5.2 ranks first in pooled Elo (1,789) from Challenge alone (\cref{fig:opencode-elo}) and GPT-6.2 Astra outscores GPT-6 Astra on 11 of 15 shared tasks (see \pubapp{sec:app-generation-results}).}

\subsection{Comparison of Multi-Round Feedback (RQ2)}
\label{sec:rq2}

\begin{figure*}[!t]
\centering
\includegraphics[width=0.98\linewidth]{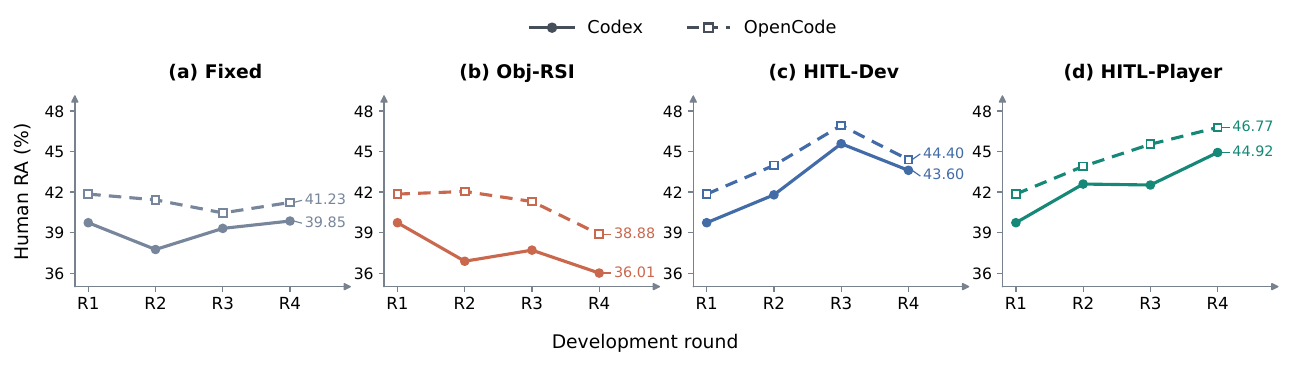}
\caption{\textbf{Human RA ratings across four development rounds on 15 paired tasks.} R1 is shared across strategies within each harness.}
\label{fig:feedback-trajectories}
\end{figure*}

\input{tab/multiround}

As seen in \cref{tab:multiround,fig:feedback-trajectories}, four strategies continue Astra across R2--R4 on the 15 challenge tasks under its two top-RA harnesses (OpenCode and Codex), branching from 30 shared R1 versions into 360 revised versions.

\paragraph{Player feedback leads R4 quality while \pubedit{Obj-RSI} lowers RA.}
At R4, HITL-Player outperforms \pubedit{Obj-RSI} by 8.92 RA points under Codex and 7.89 under OpenCode (both Holm-adjusted $p\approx0.0015$), and beats Fixed by 5.07 and 5.55 points ($p=0.0063$ and $0.0049$).
Excluding prior player-feedback providers for the same task and harness from both R4 groups leaves HITL-Player ahead of \pubedit{Obj-RSI} by 3.67 RA points under Codex and 5.01 under OpenCode.
Crucially, \pubedit{Obj-RSI} raises Auto scores by $+3.49$ and $+4.45$ points while \emph{lowering} human RA by $-3.72$ and $-2.97$ points.
HITL-Dev peaks at R3 before dipping in R4, yet records the lowest adjacent-round tier regression rates across all transitions (20.0\% and 15.6\% vs.\ 46.7\% and 33.3\% for \pubedit{Obj-RSI}), whereas HITL-Player maintains an overall upward trend across rounds to achieve the highest R4 RA.

\begin{figure*}[!t]
\centering
\includegraphics[width=\textwidth]{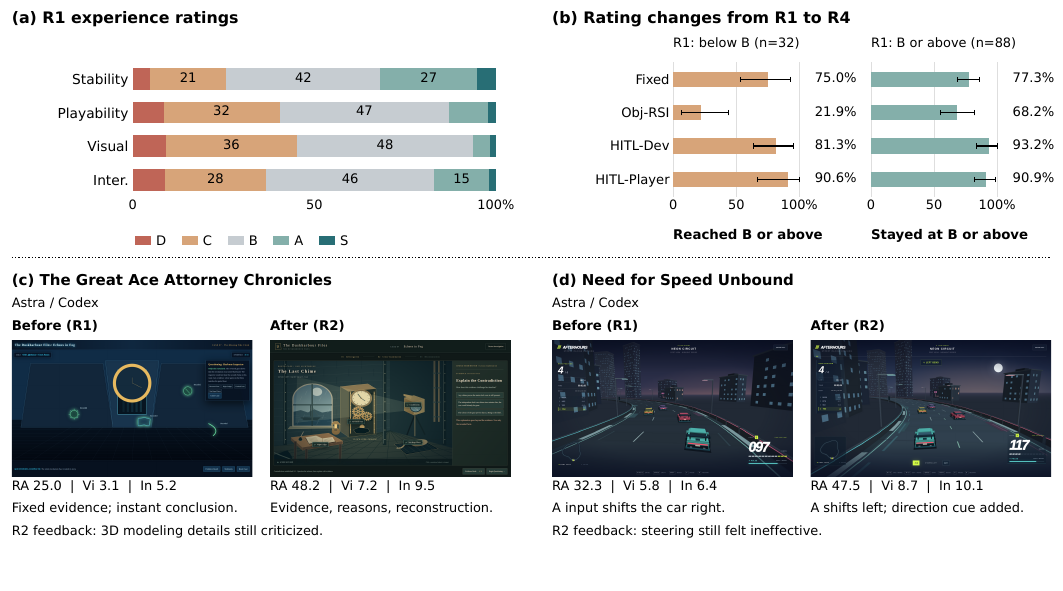}
\caption{\textbf{Demonstration of experience deficits and revision outcomes.}
(a) Task- and harness-balanced R1 tiers from 330 qualified ratings.
(b) R4 recovery means a task--harness--dimension unit with R1 median below B reaches B or above; retention means a baseline B-or-above unit remains there.
Bars and labels show percentages; $n$ denotes eligible units, and error bars are 95\% task-cluster bootstrap intervals.
(c,d) Two illustrative high-gain Astra/Codex player-feedback cases compare R1 with R2, with RA, Visual (Vi), and Inter. (In) scores (see \pubapp{sec:app-feedback-cases}).}
\label{fig:experience-issues-revision}
\end{figure*}

\input{tab/gra_main}

\paragraph{Experiential bottlenecks and dimension profiles.}
Within-task regression of RA on the four 0--25 dimensions across the 977 multi-version scenes (from the 135 tasks; $R^2_{\text{wt}}=0.865$) gives \textbf{Playability} the largest partial weight ($\beta=1.74$, 95\% CI $[1.53,1.96]$), ahead of Visual ($1.14$), Inter.\ ($0.74$), and Stability ($0.66$).
Splitting raters into disjoint halves per version (one half scoring dimensions, the other scoring RA) confirms Playability remains the primary driver ($\beta=1.01$, within-task correlation $r_{\text{wt}}=0.422$).
Thus, \emph{higher visuals rarely compensate for weak gameplay or interaction}: within tasks, high-Visual/low-Playability scenes ($n=129$) fall below the task RA mean ($-0.23$), whereas low-Visual/high-Playability ($n=133$) scores above ($+1.39$; gap $+1.63$, 95\% CI $[0.73,2.68]$; univariately, high-Inter./low-Visual scenes also outscore high-Visual/low-Inter.\ scenes at $+0.90$ vs.\ $-0.14$ because Inter.\ co-varies with Playability).
Alongside the four-dimension mean ($R^2_{\text{wt}}=0.855$), the weakest dimension $D_{\min}$ adds a positive partial slope ($\beta_{\min}=0.68$, 95\% CI $[0.32,1.03]$) and matches the four-dimension mean in disjoint-half correlation ($r_{\text{wt}}=0.428$ vs.\ $0.429$, ahead of $0.342$ for $D_{\max}$).

\Cref{tab:multiround} (\emph{Mean} rows) and \cref{fig:experience-issues-revision} compare strategy changes from the 30 shared R1 baselines, where Visual ($21/30$) and Playability ($7/30$) are lowest in $28/30$ cases (D/C: 45.3\% Visual, 40.4\% Playability):
\textbf{HITL-Dev} ($\Delta\mathrm{RA}=+3.21$) achieves the largest initial-bottleneck gain ($\Delta D_{\text{weak}}=+1.34$, with $+1.04$ Playability and $+0.86$ Visuals overall), but Codex Stability drops ($-1.05$ vs.\ $-0.08$ OpenCode), leaving non-bottleneck dimensions flat on average ($\Delta D_{\text{other}}=+0.02$).
\textbf{HITL-Player} ($\Delta\mathrm{RA}=+5.06$) lifts both the R1 bottleneck ($\Delta D_{\text{weak}}=+1.17$) \emph{and} non-bottleneck dimensions ($\Delta D_{\text{other}}=+0.69$), driven by Playability ($+1.54$) and Stability ($+1.20$) alongside Visuals ($+0.42$).
\textbf{\pubedit{Obj-RSI}} leaves bottlenecks flat ($+0.01$) and degrades $\Delta D_{\text{other}}$ ($-0.98$: Inter.\ $-0.93$, Stability $-0.91$, Playability $-0.76$).
In \mbox{\cref{fig:experience-issues-revision}}b, conditioning on R1~$<$~B includes mean reversion: Fixed recovers 24/32 sub-B units (75.0\%) while losing 20/88 competent units (retaining 77.3\%). Against this shared-R1 control, HITL-Player recovers 29/32 (90.6\%) while retaining 80/88 (90.9\%), HITL-Dev recovers 26/32 (81.3\%) and retains 82/88 (93.2\%), whereas \pubedit{Obj-RSI} recovers 7/32 (21.9\%) and retains only 60/88 (68.2\%, including just 9/22 in Playability).
Illustrative cases (\cref{fig:experience-issues-revision}c,d) show concrete player-guided deduction and steering fixes (see \pubapp{sec:app-feedback-cases}).

\subsection{GRA and Grader Agreement (RQ3)}
\label{sec:rq1-agree}

\paragraph{Programmatic checks beyond runnability.}
Across the 979 rated versions in the main study, human RA averages 39.86\% versus 84.84 for Auto, with weak rank correlation ($\rho=0.174$; $0.025$ among the 715 versions with $\mathrm{Auto}\geq80$; see \pubapp{sec:app-reliability}): Auto gates filter broken builds, whereas checklist totals weakly reflect playable experience.

\paragraph{Game Reviewer Agent (GRA) assessment.}
\label{sec:rq1-ood}
{\normalcolor Shared human replays isolate scoring by giving reviewers identical footage, while autonomous play tests their combined ability to explore a game and judge its quality from that experience. On shared replays (\cref{tab:gra-main}), \pubedit{trained GRA-35B scores align more closely with human ratings than frozen Astra's uncalibrated scores (DEV/TEST MAE 6.92/7.45 vs.\ 25.43/24.82). TEST rank correlations are 0.383 for GRA-35B and 0.381 for Astra.} In autonomous play, GRA-GRPO-9B lowers TEST MAE from 34.62 to 7.85 relative to frozen Qwen3.5-9B and raises separated-pair accuracy from 61.42\% to 69.75\%.}

\input{fig/aaa_analysis}

\FloatBarrier
\input{sec/results_extensional}

\FloatBarrier

%% file: tab/leaderboard.tex
\begin{table*}[!t]
\centering
\caption{\textbf{Single-round OpenCode results.} Pass is a percentage; Auto, RA, and $\mathrm{EC}$ range from 0 to 100. The four human EC dimension ratings range from 0 to 25: Stability (Stab.), Playability (Play.), Visual Quality (Visual), and Interactive Feedback (Inter.). Final gate failures enter as zeros, and paired differences in text use unrounded means. Bold marks the highest Auto and human scores within each tier.}
\label{tab:leaderboard}
\small
\setlength{\tabcolsep}{2.5pt}
\renewcommand{\arraystretch}{1.05}
\begin{tabular*}{\textwidth}{@{\extracolsep{\fill}}llrrrlrrrrr@{}}
\toprule
\tabhead{Evaluation tier} & Model & $n$ & Pass & Auto & RA & $\mathrm{EC}$ & Stab. & Play. & Visual & Inter. \\
\midrule
Easy & GPT-6 Astra & 30 & 100.0 & 81.95 & \textbf{46.61} & \textbf{36.43} & \textbf{10.57} & \textbf{8.57} & \textbf{8.43} & \textbf{8.85} \\
Easy & Opus 5.5 & 30 & 100.0 & \textbf{86.75} & 40.94 & 30.32 & 8.61 & 8.06 & 6.54 & 7.11 \\
Easy & Fable 5.1 & 30 & 100.0 & 73.62 & 37.46 & 28.00 & 8.33 & 6.87 & 5.79 & 7.01 \\
Easy & Opus 5 & 30 & 76.7 & 52.98 & 29.17 & 22.54 & 6.13 & 5.64 & 5.60 & 5.17 \\
Easy & Kimi K3 & 30 & 90.0 & 67.33 & 28.50 & 20.48 & 6.46 & 5.16 & 3.96 & 4.90 \\
Easy & GPT-5.6 Sol & 30 & 96.7 & 71.87 & 26.85 & 20.77 & 7.10 & 4.83 & 3.96 & 4.88 \\
Easy & GLM-5.3 & 30 & 83.3 & 60.43 & 24.45 & 17.67 & 5.39 & 4.25 & 3.67 & 4.35 \\
Easy & DeepSeek V4-Flash & 30 & 70.0 & 50.98 & 13.66 & 10.03 & 3.53 & 2.28 & 1.98 & 2.25 \\
\midrule
Medium & GPT-6 Astra & 40 & 100.0 & 77.16 & \textbf{48.21} & \textbf{39.02} & \textbf{11.37} & \textbf{9.14} & \textbf{8.86} & \textbf{9.65} \\
Medium & Opus 5.5 & 40 & 97.5 & \textbf{80.63} & 38.99 & 29.25 & 8.40 & 7.61 & 6.32 & 6.93 \\
Medium & Fable 5.1 & 40 & 100.0 & 76.89 & 37.64 & 28.64 & 8.63 & 7.26 & 5.92 & 6.82 \\
Medium & Opus 5 & 40 & 100.0 & 75.45 & 35.20 & 25.87 & 7.08 & 6.44 & 6.10 & 6.25 \\
\midrule
Hard & GPT-6 Astra & 50 & 100.0 & \textbf{87.61} & \textbf{42.71} & \textbf{33.24} & \textbf{9.40} & 7.91 & \textbf{7.98} & \textbf{7.95} \\
Hard & Opus 5.5 & 50 & 96.0 & 83.68 & 40.37 & 31.08 & 8.89 & \textbf{8.11} & 6.69 & 7.40 \\
\bottomrule
\end{tabular*}
\end{table*}

%% file: fig/harness_main.tex
\begin{figure*}[!t]
\centering
\includegraphics[width=\textwidth]{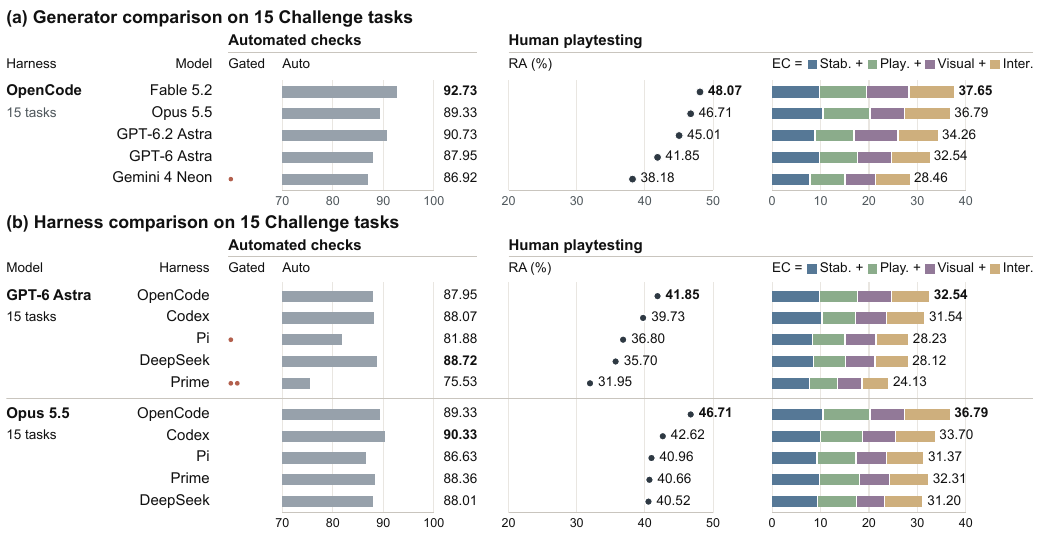}
\caption{\textbf{R1 on 15 Challenge tasks.} (a) Five OpenCode generators, three added in follow-up (45 candidates, 44 playable, 640 qualified ratings). (b) Two generators across five harnesses. DeepSeek denotes its CLI. Dots count gate failures. Metrics follow \cref{tab:leaderboard}.}
\label{tab:harness}
\end{figure*}

%% file: sec/5_setup.tex
\paragraph{Experimental setup.}
\label{sec:setup}
OpenCode supports the model comparisons across Easy, Medium, and Hard tiers; the separate 15-task challenge compares GPT-6 Astra and Opus~5.5 across five harnesses, then continues Astra under Codex and OpenCode across four revision rounds. Unless stated otherwise, results use qualified human ratings and candidate-level comparisons include final gate failures as zeros.

%% file: tab/multiround.tex
\begin{table}[!htb]
\centering
\caption{\textbf{Multi-round R4$-$R1 gains and bottleneck repair on 15 paired tasks per harness.} RA and Auto use 0--100 scales; the four EC dimensions use 0--25. $\Delta D_{\text{weak}}$ tracks each baseline's lowest-scoring R1 dimension, $\Delta D_{\text{other}}$ averages the remaining three, and \emph{Mean} averages Codex and OpenCode.}
\label{tab:multiround}
\makeatletter\@ifclassloaded{lumia}{\footnotesize\setlength{\tabcolsep}{4pt}}{\fontsize{7}{8.4}\selectfont\setlength{\tabcolsep}{0.6pt}}\makeatother
\renewcommand{\arraystretch}{1.05}
\begin{tabular*}{\columnwidth}{@{\extracolsep{\fill}}llrrrrrrrr@{}}
\toprule
 & Strategy & RA & Auto & Stab. & Play. & Vis. & Inter. & $\Delta D_{\mathrm{weak}}$ & $\Delta D_{\mathrm{other}}$ \\
\midrule
\multirow{4}{*}{\rotatebox{90}{Codex}} & Fixed & $+0.13$ & $+1.79$ & $-1.34$ & $+0.46$ & $+0.24$ & $-1.08$ & $+0.61$ & $-0.78$ \\
 & \pubedit{Obj-RSI} & $-3.72$ & $+3.49$ & $-1.59$ & $-0.71$ & $-0.33$ & $-1.54$ & $-0.10$ & $-1.36$ \\
 & HITL-Dev & $+3.88$ & $+1.51$ & $-1.05$ & $+1.07$ & $\mathbf{+1.25}$ & $\mathbf{+0.12}$ & $\mathbf{+1.83}$ & $-0.15$ \\
 & HITL-Player & $\mathbf{+5.19}$ & $\mathbf{+3.55}$ & $\mathbf{+0.90}$ & $\mathbf{+1.83}$ & $+0.38$ & $-0.03$ & $+1.13$ & $\mathbf{+0.65}$ \\
\midrule
\multirow{4}{*}{\rotatebox{90}{OpenCode}} & Fixed & $-0.62$ & $+2.81$ & $-0.95$ & $-0.51$ & $-0.15$ & $-0.51$ & $+0.35$ & $-0.83$ \\
 & \pubedit{Obj-RSI} & $-2.97$ & $\mathbf{+4.45}$ & $-0.22$ & $-0.82$ & $-0.33$ & $-0.32$ & $+0.12$ & $-0.61$ \\
 & HITL-Dev & $+2.55$ & $+3.51$ & $-0.08$ & $+1.00$ & $\mathbf{+0.46}$ & $+0.05$ & $+0.86$ & $+0.19$ \\
 & HITL-Player & $\mathbf{+4.93}$ & $+2.30$ & $\mathbf{+1.50}$ & $\mathbf{+1.25}$ & $\mathbf{+0.46}$ & $\mathbf{+0.20}$ & $\mathbf{+1.21}$ & $\mathbf{+0.73}$ \\
\midrule
\multirow{4}{*}{\rotatebox{90}{Mean}} & Fixed & $-0.25$ & $+2.30$ & $-1.15$ & $-0.02$ & $+0.04$ & $-0.80$ & $+0.48$ & $-0.80$ \\
 & \pubedit{Obj-RSI} & $-3.35$ & $\mathbf{+3.97}$ & $-0.91$ & $-0.76$ & $-0.33$ & $-0.93$ & $+0.01$ & $-0.98$ \\
 & HITL-Dev & $+3.21$ & $+2.51$ & $-0.57$ & $+1.04$ & $\mathbf{+0.86}$ & $\mathbf{+0.09}$ & $\mathbf{+1.34}$ & $+0.02$ \\
 & HITL-Player & $\mathbf{+5.06}$ & $+2.93$ & $\mathbf{+1.20}$ & $\mathbf{+1.54}$ & $+0.42$ & $+0.08$ & $+1.17$ & $\mathbf{+0.69}$ \\
\bottomrule
\end{tabular*}
\end{table}

%% file: tab/gra_main.tex
\begin{table*}[!t]
\centering
\captionsetup{skip=3pt}
\caption{\textbf{Reviewer agreement with human ratings.} Each version receives equal weight. Sep. is accuracy on within-task pairs with statistically separated human ratings, counting prediction ties as half correct.}
\label{tab:gra-main}
\footnotesize
\setlength{\tabcolsep}{4pt}
\renewcommand{\arraystretch}{1.0}
\begin{tabular*}{\textwidth}{@{\extracolsep{\fill}}l*{6}{r}@{}}
\toprule
 & \multicolumn{3}{c}{\textbf{DEV}} & \multicolumn{3}{c}{\textbf{TEST}} \\
\cmidrule(lr){2-4}\cmidrule(l){5-7}
\textbf{Reviewer} & \textbf{\boldmath MAE $\downarrow$} & \textbf{\boldmath Spearman $\rho\uparrow$} & \textbf{\boldmath Sep. (\%) $\uparrow$} & \textbf{\boldmath MAE $\downarrow$} & \textbf{\boldmath Spearman $\rho\uparrow$} & \textbf{\boldmath Sep. (\%) $\uparrow$} \\
\midrule
\multicolumn{7}{@{}l}{\textbf{Shared human replays}}\\[1pt]
Qwen3.5-35B-A3B, frozen & 29.68 & 0.291 & 72.10 & 30.47 & 0.380 & 77.16 \\
GPT-6 Astra, frozen & 25.43 & 0.344 & 82.01 & 24.82 & 0.381 & 76.34 \\
GRA-35B & \textbf{6.92} & \textbf{0.401} & \textbf{83.26} & \textbf{7.45} & \textbf{0.383} & \textbf{77.42} \\
\addlinespace[2pt]\midrule
\multicolumn{7}{@{}l}{\textbf{Autonomous play and self-scoring}}\\[1pt]
Qwen3.5-9B, frozen & 33.75 & 0.092 & 52.15 & 34.62 & 0.268 & 61.42 \\
Qwen3.5-35B-A3B, frozen & 34.33 & 0.137 & 64.81 & 35.85 & 0.218 & 54.01 \\
GRA-GRPO-9B & \textbf{7.58} & \textbf{0.305} & \textbf{71.67} & \textbf{7.85} & \textbf{0.322} & \textbf{69.75} \\
\bottomrule
\end{tabular*}\par\vspace{2pt}
\end{table*}

%% file: fig/aaa_analysis.tex
\begin{figure}[!t]
\centering
\makeatletter\@ifclassloaded{lumia}{\includegraphics[width=0.72\linewidth]{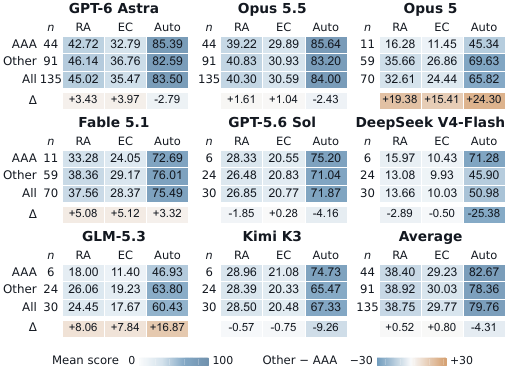}}{\includegraphics[width=\columnwidth]{fig/aaa_overall_score_heatmap.pdf}}\makeatother
\caption{\textbf{AAA-referenced versus other tasks.} \emph{Average} summarizes the eight main-study models. For each task, we average versions within each evaluated model (gate failures score zero), then average those models equally. Each task has equal weight within its group. $n$ counts tasks and $\Delta$ gives Other$-$AAA in points.}
\label{fig:aaa-analysis}
\end{figure}

%% file: sec/results_extensional.tex
\subsection{\pubedit{How Closely Do Generated Scenes Match AAA References?}}
\label{sec:extensional-analysis}

AAA references account for 6/30 Easy, 5/40 Medium, 27/50 Hard, and 6/15 Challenge tasks (\cref{fig:aaa-analysis}). Across all 135 tasks, the Other$-$AAA gap is 0.52 points for RA (95\% CI $[-1.94,3.14]$) and $-4.31$ for Auto. On Easy, GPT-5.6 Sol scores 28.33 RA on AAA and 26.48 on Other. \pubedit{Within the same model and evaluation tier, human ratings show a \textbf{3.31-point RA deficit on AAA references}} ($[1.23,5.49]$). The Auto gap is near zero at $-0.01$ points ($[-2.56,2.49]$), indicating that the pooled Auto advantage for AAA tasks largely reflects the mix of tiers and models.

%% file: sec/8_conclusion.tex
\section{Conclusion}
\label{sec:conclusion}
\ours introduces a 1,001-game pool (a 135-task open set and a dynamic closed set) uniting play-before-judge ratings, multi-round revision, and automated review. GPT-6 Astra leads three OpenCode tiers while Opus~5.5 leads the 15-task five-harness challenge. Human feedback outperforms \pubedit{Obj-RSI} at R4 across playability bottlenecks and overall experience, whereas \pubedit{Obj-RSI} lowers RA. \pubedit{Training aligns GRA scores more closely with human judgments, reducing TEST mean absolute error} from 30.47 to 7.45 for 35B replay scoring and from 34.62 to 7.85 for 9B autonomous play and self-scoring.

\paragraph{Limitations and ethics.}
\label{sec:limitations}
We evaluate scoped Three.js scenes, with revision on 15 challenge tasks and GRA assessed across 25 task families per split, leaving full-length game production and broader engine transfer to future work. Prompts require original code and assets using commercial titles only as references.

%% file: sec/9_llm_usage.tex


%% file: sec/8_authors.tex
\section{Authors}
\label{sec:authors}
\begingroup
\small\raggedright
\mbox{Pengfei~Zhou$^{1,*}$}
\quad
\mbox{Yixing~Ma$^{1,2,3,*}$}
\quad
\mbox{Weidong~Tang$^{1,4,*}$}
\quad
\mbox{Hexin~Wang$^{1,*}$}
\quad
\mbox{Shuo~Liang$^{1,5}$}
\quad
\mbox{Yi~Zhang$^{1}$}
\quad
\mbox{Ruizhi~Qiu$^{6}$}
\quad
\mbox{Meng~Luo$^{2}$}
\quad
\mbox{Yanlin~Li$^{2}$}
\quad
\mbox{Hongzhan~Lin$^{2}$}
\quad
\mbox{Bowen~Li$^{1}$}
\quad
\mbox{Jiancheng~Pan$^{7}$}
\quad
\mbox{Zhongjie~Lao$^{1,2}$}
\quad
\mbox{Qinglong~Cao$^{8}$}
\quad
\mbox{Zhengfeiyang~Zhang$^{1,9}$}
\quad
\mbox{Haoyuan~Chen$^{1,10}$}
\quad
\mbox{Bin~Xu$^{1}$}
\quad
\mbox{Yingxuan~Ren$^{2}$}
\quad
\mbox{Xiaoda~Yang$^{11}$}
\quad
\mbox{Yuhui~Xie$^{12}$}
\quad
\mbox{Jianye~Pang$^{12}$}
\quad
\mbox{Siyao~Wang$^{1}$}
\quad
\mbox{Chongbin~Yu$^{5}$}
\quad
\mbox{Siyuan~Ma$^{13}$}
\quad
\mbox{Jiachen~Luo$^{14}$}
\quad
\mbox{Jiaqi~Lang$^{2}$}
\quad
\mbox{Run~Liu$^{15}$}
\quad
\mbox{Longtao~Wu$^{6}$}
\quad
\mbox{Shusheng~Li$^{5}$}
\quad
\mbox{Zeyu~Yan$^{11}$}
\quad
\mbox{Da~Peng$^{12}$}
\quad
\mbox{Xiao~Cui$^{16}$}
\quad
\mbox{Zhaokai~Wang$^{15}$}
\quad
\mbox{Mingyang~Wang$^{5}$}
\quad
\mbox{Suhan~Duan$^{5}$}
\quad
\mbox{Anqi~Li$^{6}$}
\quad
\mbox{Jiangming~Wang$^{6}$}
\quad
\mbox{Yixu~Huang$^{17}$}
\quad
\mbox{Lekang~Jiang$^{18}$}
\quad
\mbox{Yaxi~Qu$^{19}$}
\quad
\mbox{Wentong~Zhao$^{5}$}
\quad
\mbox{Chunlin~Liu$^{6}$}
\quad
\mbox{Mo~Zhang$^{20}$}
\quad
\mbox{Shengrui~Han$^{21}$}
\quad
\mbox{Xinfeng~Yang$^{17}$}
\quad
\mbox{Ruilin~Tang$^{22}$}
\quad
\mbox{Zhenglin~Wan$^{2}$}
\quad
\mbox{Yeying~Jin$^{2}$}
\quad
\mbox{Kaipeng~Zhang$^{23}$}
\quad
\mbox{Wangbo~Zhao$^{4,\dagger}$}
\quad
\mbox{Yang~You$^{2,\dagger}$}\par\smallskip
\noindent $^*$Equal contribution.\quad $^\dagger$Corresponding authors.\par
{\footnotesize\href{mailto:zpf4wp@outlook.com}{zpf4wp@outlook.com}\quad\href{mailto:wangbo.zhao96@gmail.com}{wangbo.zhao96@gmail.com}\quad\href{mailto:yangyou@nus.edu.sg}{yangyou@nus.edu.sg}\par}
\smallskip
\noindent\textbf{Affiliations:}\par
{\footnotesize\renewcommand{\arraystretch}{1.12}
\begin{tabularx}{\linewidth}{@{}r@{\hspace{0.7em}}X@{}}
1 & InfRec, Cardinal AI Lab \\
2 & National University of Singapore \\
3 & University of California, Berkeley \\
4 & The Hong Kong University of Science and Technology \\
5 & Xidian University \\
6 & Independent Researcher \\
7 & Tsinghua University \\
8 & Shanghai AI Laboratory \\
9 & The University of Tokyo \\
10 & University of Chinese Academy of Sciences \\
11 & Zhejiang University \\
12 & Xi'an Jiaotong University \\
13 & The Chinese University of Hong Kong \\
14 & Technical University of Munich \\
15 & Shanghai Jiao Tong University \\
16 & University of Science and Technology of China \\
17 & Fudan University \\
18 & University of Cambridge \\
19 & University of Nottingham \\
20 & Shandong University \\
21 & Southeast University \\
22 & Hunan University \\
23 & Shanghai Innovation Institute \\
\end{tabularx}}
\endgroup

%% file: sec/X_suppl.tex
\makeatletter
\@ifclassloaded{lumia}{%
  \newcommand{\appsplitright}{}%
}{%
  \clearpage
  \setcounter{page}{1}%
  \renewcommand{\thepage}{S\arabic{page}}%
  \def\appstriptitle#1.~#2\appstripend{#2}%
  \newcommand{\appentryname}{\etoclink{\expandafter\appstriptitle\etocthename\appstripend}}%
  \newcommand{\appcontents}{%
    \begingroup\normalsize\raggedright\setlength{\parindent}{0pt}%
    \textcolor{black!45}{\rule{\textwidth}{0.4pt}}\par\vspace{5pt}%
    {\large\textbf{Contents}}\par\vspace{2pt}%
    \etocsettocstyle{}{}%
    \etocsetstyle{section}{}{}%
      {\par\addvspace{3pt}\small\noindent
       \makebox[1.45em][l]{\textbf{\textcolor{rqaccent}{\etocthenumber}}}%
       \textbf{\appentryname}\nobreak\leaders\hbox to 0.6em{\hss.\hss}\hfill\nobreak\makebox[2.1em][r]{\etocthepage}\par}%
      {}%
    \etocsetstyle{subsection}{}{}%
      {\par\footnotesize\noindent\hspace*{1.45em}\makebox[2.05em][l]{\textcolor{black!62}{\etocthenumber}}%
       \appentryname\nobreak\hfill\nobreak\makebox[2.1em][r]{\textcolor{black!62}{\etocthepage}}\par}%
      {}%
    \hypersetup{linkcolor=black}%
    \begin{minipage}[t]{0.475\textwidth}
      \etocsettagdepth{mainmatter}{none}\etocsettagdepth{apxleft}{subsection}\etocsettagdepth{apxright}{none}\etocsetnexttocdepth{subsection}%
      \tableofcontents
    \end{minipage}\hfill
    \begin{minipage}[t]{0.475\textwidth}
      \etocsettagdepth{mainmatter}{none}\etocsettagdepth{apxleft}{none}\etocsettagdepth{apxright}{subsection}\etocsetnexttocdepth{subsection}%
      \tableofcontents
    \end{minipage}\par\vspace{5pt}%
    \textcolor{black!45}{\rule{\textwidth}{0.4pt}}\par
    \endgroup}%
  \twocolumn[{%
    \centering\Large\textbf{\thetitle}\\
    \vspace{0.5em}Supplementary Material\\
    \vspace{0.8em}%
    \appcontents
    \vspace{0.9em}%
  }]%
  \appendix
  \etocdepthtag.toc{apxleft}%
  \newcommand{\appsplitright}{\etocdepthtag.toc{apxright}}%
}
\setlength{\@fptop}{0pt}
\setlength{\@dblfptop}{0pt}
\setlength{\@dblfpsep}{10pt}
\setlength{\@dblfpbot}{0pt plus 1fil}
\makeatother

\crefalias{section}{appendix}
\counterwithin*{table}{section}
\renewcommand{\thetable}{\thesection\arabic{table}}
\renewcommand{\theHtable}{appendix.\thesection.\arabic{table}}

\renewcommand{\topfraction}{0.9}
\renewcommand{\dbltopfraction}{0.9}
\renewcommand{\textfraction}{0.07}
\renewcommand{\floatpagefraction}{0.8}
\renewcommand{\dblfloatpagefraction}{0.8}

\definecolor{appheadbg}{HTML}{E8EDF4}
\definecolor{appgroupbg}{HTML}{F3F5F8}
\definecolor{appzebrabg}{HTML}{F8F9FB}
\definecolor{appbestbg}{HTML}{D9E6F5}
\definecolor{apprule}{HTML}{9AA6B6}
\newcolumntype{R}{>{\raggedleft\arraybackslash}X}
\newcolumntype{L}{>{\raggedright\arraybackslash}X}
\newcolumntype{C}{>{\centering\arraybackslash}X}
\newcommand{\apptab}{\footnotesize\setlength{\tabcolsep}{5pt}\renewcommand{\arraystretch}{1.18}%
  \setlength{\aboverulesep}{0pt}\setlength{\belowrulesep}{0pt}\arrayrulecolor{apprule}}
\newcommand{\apptop}{\specialrule{0.6pt}{0pt}{0pt}}
\newcommand{\appmid}{\specialrule{0.4pt}{0pt}{0pt}}
\newcommand{\appbot}{\specialrule{0.6pt}{0pt}{0pt}}
\newcommand{\apphead}{\rowcolor{appheadbg}}
\newcommand{\appzebra}{\rowcolor{appzebrabg}}
\newcommand{\appgroup}[2]{\rowcolor{appgroupbg}\multicolumn{#1}{@{\hspace{5pt}}l}{\bfseries #2}}
\newcommand{\appgroupd}[3]{\rowcolor{appgroupbg}\multicolumn{#1}{@{\hspace{5pt}}l}{\textbf{#2}\textcolor{black!55}{~\textperiodcentered~#3}}}
\newcommand{\appbest}[1]{{\setlength{\fboxsep}{1.4pt}\colorbox{appbestbg}{\bfseries\boldmath #1}}\hspace{-1.4pt}}
\newcommand{\appci}[1]{\textcolor{black!55}{[#1]}}
\newcommand{\appcis}[1]{{\scriptsize\appci{#1}}}
\newcommand{\appmuted}[1]{\textcolor{black!55}{#1}}

\makeatletter
\@ifclassloaded{lumia}{%
  \titleformat{\section}{\large\bfseries\headingfont}{\pubedit{\thesection.}}{0.5em}{#1}%
  \titleformat{\subsection}{\color{reportaccent}\bfseries}{\pubedit{\thesubsection.}}{0.5em}{#1}%
}{}%
\makeatother

\section{\pubedit{Supplementary Implementation Details}}
\label{sec:app-implementation}\label{sec:app-protocol-benchmark}

\subsection{\pubedit{Task Pool and Generation Prompts}}
\label{sec:app-pool}\label{sec:app-prompt}

The 1,001 titles span 15 primary genres and 42 subgenres (\cref{tab:app-pool}). After generating initial builds with Gemini~3.7 Flash under OpenCode and scoring them via Auto (Codex with GPT-6 Astra), we randomly sample 120 tasks across the three \pubedit{EEE} tiers (30 Easy, 40 Medium, and 50 Hard) plus 15 Challenge tasks to form the fixed \textbf{135-task open set} used for all generator comparisons. The remaining 866 titles stay withheld from public leaderboards in a \textbf{closed set} that rotates tasks within genre strata across evaluation waves; 120 internal pilot tasks from this withheld pool also supply playtest trajectories strictly for GRA ({\pubedit{\cref{sec:app-gra}}}).

\paragraph{System and R1 prompts.}
Agents run under the system instruction:
\begin{quote}
\small\raggedright
You are generating a game for an automated benchmark. Implement the requested game directly inside the current working directory. Work autonomously, do not ask the user questions, and keep all created files inside the workspace. The framework only collects generation artifacts; do not attempt to score or evaluate the game.
\end{quote}
The R1 user prompt specifies:
\begin{quote}
\small\raggedright
Use Three.js to build a high-quality, original $\langle$\textit{subgenre}$\rangle$ game; borrow only the core gameplay mechanics of $\langle$\textit{reference game}$\rangle$, and generate high-quality 3D visuals. Characters, weapons\slash items, abilities, levels, art, audio, UI and code must be original; reusing existing names, character likenesses, character models or assets is strictly prohibited, and names must be clearly different.
\end{quote}

\begin{table*}[!htbp]
\centering
\caption{\textbf{Task pool.} Reference tasks by primary genre across the 1,001-game pool.}
\label{tab:app-pool}
\apptab
\begin{tabularx}{\linewidth}{@{\hspace{5pt}}lRR@{\hspace{16pt}}lRR@{\hspace{5pt}}}
\apptop
\apphead {\bfseries\boldmath Primary genre} & {\bfseries\boldmath Tasks} & {\bfseries\boldmath Share (\%)} & {\bfseries\boldmath Primary genre} & {\bfseries\boldmath Tasks} & {\bfseries\boldmath Share (\%)} \\
\appmid
Action & 166 & 16.6 & Racing & 39 & 3.9 \\
\appzebra Simulation & 116 & 11.6 & Sandbox \& survival & 39 & 3.9 \\
Casual & 103 & 10.3 & Horror & 39 & 3.9 \\
\appzebra Role-playing & 94 & 9.4 & Puzzle & 34 & 3.4 \\
Strategy & 91 & 9.1 & Music \& rhythm & 26 & 2.6 \\
\appzebra Shooter & 84 & 8.4 & Experimental & 13 & 1.3 \\
Sports & 78 & 7.8 & Tabletop & 13 & 1.3 \\
\appzebra Adventure & 66 & 6.6 & {\bfseries\boldmath Total} & {\bfseries\boldmath 1,001} & {\bfseries\boldmath 100.0} \\
\appbot
\end{tabularx}
\end{table*}

\begin{table*}[!htbp]
\centering
\caption{\textbf{R2--R4 continuation prompt templates across all four feedback strategies.} Brackets denote artifact-specific inputs; each branch continues from the preceding runnable workspace.}
\label{tab:app-fixed}
\apptab
\begin{tabularx}{\linewidth}{@{\hspace{5pt}}>{\raggedright\arraybackslash}p{2.3cm}L@{\hspace{5pt}}}
\apptop
\apphead {\bfseries Strategy} & {\bfseries Continuation prompt structure (rounds R2--R4)} \\
\appmid
{\bfseries Fixed} & \textbf{R2:} Extend $\langle$\textit{specified mechanic}$\rangle$ and replace placeholder assets. \textbf{R3:} Browser-test $\langle$\textit{edge cases}$\rangle$, fix issues directly, and replay. \textbf{R4:} Preserve core rules, numerical parameters, and state machines; polish $\langle$\textit{visuals and feedback}$\rangle$ and optimize frame rate for low-end hardware. \\
\appzebra {\bfseries \pubedit{Obj-RSI}} & Continue from the preceding round's runnable build; preserve working features, fix reported issues, and verify in a browser. Supplies $\langle$\textit{R1 brief}$\rangle$ plus structured JSON fields \texttt{rubric}, \texttt{score}, \texttt{gate}, and \texttt{issues} (deficient or unverified checks with \texttt{id}, \texttt{name}, \texttt{status}, \texttt{score}, \texttt{maxScore}, \texttt{expected}, \texttt{observed}, \texttt{evidence}). \\
{\bfseries HITL-Dev} ($n\ge 3$) & Passes the developer's concise playtest critique note directly to revise the preceding build (e.g., \emph{``The level is too simple; improve core gameplay and 3D models.''}). \\
\appzebra {\bfseries HITL-Player} & Supplies player ratings on the preceding build ($\langle$\textit{mean overall}$\rangle/100$ and 0--4 means for Playability $\langle P\rangle$, Stability $\langle S\rangle$, Visuals $\langle V\rangle$, and Interaction/Feedback $\langle F\rangle$) together with $\langle$\textit{summarized player comments}$\rangle$, instructing the agent to refine the build and play-test in browser. \\
\appbot
\end{tabularx}
\end{table*}

\subsection{\pubedit{Human Rubric and Automated Checks}}
\label{sec:app-rubric}\label{sec:app-obj}

Annotators first review the reference game's description and gameplay video outside the playtest timer, play the generated scene, and report overall Reference Attainment $P\in[0,100]$ (5\% steps: 0--49\% clearly below similar games, 50--74\% recognizable playable loop with clear gaps, 75--100\% close to reference within scene scope) alongside S--D grades for four dimensions (\cref{tab:app-rubric}), whose sum defines the Experience Composite ($\mathrm{EC}$).

\begin{table*}[!htbp]
\centering
\caption{\textbf{Tier criteria for the four human rating dimensions.} Grey numbers give point values; summing the four dimensions defines the Experience Composite ($\mathrm{EC}\in[0,100]$).}
\label{tab:app-rubric}
\apptab
\begin{tabularx}{\linewidth}{@{\hspace{5pt}}>{\raggedright\arraybackslash}p{2.35cm}*{5}{L}@{\hspace{5pt}}}
\apptop
\apphead {\bfseries Dimension} & {\bfseries S}~\appmuted{(25)} & {\bfseries A}~\appmuted{(16.7)} & {\bfseries B}~\appmuted{(8.3)} & {\bfseries C}~\appmuted{(4.2)} & {\bfseries D}~\appmuted{(0)} \\
\appmid
{\bfseries Stability} & Consistently smooth; no blocking issues & Reliable overall; only minor issues & Main flow completable, with noticeable anomalies & Starts, but frequent issues disrupt play & Fails to start, crashes, or breaks main flow \\
\appzebra {\bfseries Playability} & Complete core gameplay; natural controls close to reference & Core loop functional and smooth; minor feel/rule issues & Main gameplay playable, with clear control shortcomings & Basic functions only; incomplete loop & No effective gameplay or uncontrollable \\
{\bfseries Visual Quality} & Visuals reach similar games with high coherence & Highly complete; minor details lacking & Style and scene recognizable but clearly simplified & Basic layout and primitive elements only & Severe omissions or rendering failures \\
\appzebra {\bfseries Interactive Feedback} & Timely, clear, layered feedback at reference level & Main feedback complete and clear; minor omissions & Core actions give weak or inconsistent feedback & Sparse, delayed, or unclear feedback & Key feedback missing/wrong; state unclear \\
\appbot
\end{tabularx}
\end{table*}

For Auto (\cref{tab:app-obj}), a Codex agent running GPT-6 Astra with Playwright checks six sequential gates: reachable page, initialized runtime, visible canvas, nonblank rendering, reachable play, and no blocking runtime error. Any gate failure scores zero; passing builds receive $100\times\text{earned}/\text{applicable}$ points (rounded to one decimal, with integer partial credit and unobserved items scored zero).

\begin{table*}[!htbp]
\centering
\caption{\textbf{All 41 Auto checks across ten categories} (100 points total; parentheses give item points, and $^*$ marks the 15 conditional items omitted from the denominator when inapplicable to a game's design).}
\label{tab:app-obj}
\apptab
\begin{tabularx}{\linewidth}{@{\hspace{5pt}}lL@{\hspace{5pt}}}
\apptop
\apphead {\bfseries Category (pts)} & {\bfseries Checks in rubric order} \\
\appmid
Startup (15) & Page loads (3), runtime initializes (4), visible canvas created (3), core assets load (2), no blocking startup error (3). \\
\appzebra Rendering (15) & Nonblank canvas (5), visible scene (3), continuous animation ($3^*$), usable camera and lighting (2), viewport resizes (2). \\
Entry (10) & Exits loading screen (3), start control works ($3^*$), transitions into play (2), no blocking overlay (2). \\
\appzebra Keyboard (12) & Primary keys respond ($5^*$), distinct key effects ($3^*$), key release stops input ($2^*$), focus works ($2^*$). \\
Mouse/touch (8) & Click response ($3^*$), move/drag/wheel ($2^*$), touch controls ($2^*$), accurate hit coordinates ($1^*$). \\
\appzebra Basic play (15) & Player/camera controllable (5), sustained play for 10--30\,s (4), rule-consistent state updates (3), no immediate deadlock (3). \\
Rule feedback (10) & Core action outcome (4), score/progress updates ($3^*$), rule constraints hold ($1^*$), feedback matches state (2). \\
\appzebra Game loop (8) & Start--play--end flow (2), win/loss/completion trigger ($2^*$), recognizable end state ($2^*$), clean restart ($2^*$). \\
Stability (4) & Error-free play (1), survives 10 rapid inputs (1), repeat play/reload works (1), recovers after focus switch (1). \\
\appzebra Performance (3) & Interactive within 10\,s (1), mean $\geq30$ FPS over $\geq3$\,s with no 1\,s $<20$ FPS (1), no severe stall or heap growth (1). \\
\appbot
\end{tabularx}
\end{table*}

\subsection{\pubedit{Annotator Qualification, Blind Assignment, and Outlier Screening}}
\label{sec:app-annotators}

From over 1,000 volunteers recruited through crowdsourcing, we invited applicants with relevant gameplay experience to an interview that included a trial playtest. We admitted candidates who could apply the rubric and support their ratings with specific observations from play, and 63 qualified annotators contributed ratings to the study. Playtesters evaluate all builds double-blind to generator model, harness, revision round (R1--R4), and feedback strategy, with candidate versions of each task randomly assigned and presented in randomized order within evaluation waves.

After deduplicating to each rater's latest submission per version (12,136 raw sessions across 979 playable versions), our operational screen fits a regularized MAP Student-$t$ model with wave intercepts $\mu_w$, version effects $a_i\sim\mathcal N(0,30^2)$, rater offsets $b_u\sim\mathcal N(0,15^2)$, rater-by-wave interactions $c_{uw}\sim\mathcal N(0,10^2)$, and heavy-tailed residuals $\varepsilon_{uiw}\sim t_{\nu=4}(0,20)$:
\begin{equation}
y_{uiw}=\mu_w+a_i+b_u+c_{uw}+\varepsilon_{uiw}.
\label{eq:app-student-t-screen}
\end{equation}
Using median predictions $\hat y_{uiw}$ from repeated five-fold within-version out-of-fold fits, submissions on versions with $\ge 3$ raters exceeding $|y_{uiw}-\hat y_{uiw}|>20\,t_4^{-1}(0.90)=30.664$ points are removed.

Leaderboard RA then averages the retained raw ratings $p_{iu}$ arithmetically within each version (without subtracting rater offsets $b_u$) and weights tasks equally within each configuration.

\subsection{\pubedit{Game Reviewer Agent Details}}
\label{sec:app-gra}

\textbf{Reviewer data.} The 135 open-set tasks contribute 979 rated builds. Another 120 withheld pilot tasks contribute 234 builds rated under the same blind protocol. We split these 255 task families into 205 TRAIN, 25 DEV, and 25 TEST.

Whereas leaderboard RA measures the unadjusted mean score assigned by playtesters, GRA predicts typical version quality after removing individual rater leniency. Let $p_{iu}$ denote qualified raw rating $u$ on version $i$. Using TRAIN versions only, we fit an additive Huber regression with version intercepts and $L_2$-regularized rater offsets $b_u$ (regularization strength chosen via 3-fold task-grouped cross-validation within TRAIN). Freezing $\hat b_u$ ($\hat b_u=0$ for raters unseen in TRAIN), the target and evaluation label on all splits is
\begin{equation}
y_i^{\mathrm{GRA}} = \operatorname{clip}_{[0,100]}\!\left(\operatorname{median}_{u}\bigl(p_{iu}-\hat b_u\bigr)\right).
\label{eq:app-gra-target}
\end{equation}

\textbf{Reviewer agreement.} Shared-replay reviewers score the same 16 frames in one call per replay. In autonomous play, each reviewer plays and scores its own trajectory. MAE and Spearman's $\rho$ weight each version equally. Sep. counts a within-task pair when its human targets differ by more than $1.96\sqrt{\mathrm{SE}_i^2+\mathrm{SE}_j^2}$. Correct orderings earn one point and predicted ties earn half.

\clearpage

\appsplitright

\section{\pubedit{Supplementary Experimental Results}}
\label{sec:app-results}

\subsection{\pubedit{Single-Round Generation and OpenCode Elo}}
\label{sec:app-generation}\label{sec:app-generation-results}\label{sec:app-population}

\Cref{tab:scale} details the five open-set evaluation groups across 30 Easy, 40 Medium, 50 Hard, and 15 Challenge tasks, comprising 1,010 candidates (979 playable builds and 31 gate failures); under \pubedit{EEE} with OpenCode, the top four Easy generators by RA advance to Medium and the top two advance to Hard. \Cref{tab:app-leaderboard-full,tab:app-harness-full} provide the full single-round model and harness results with 95\% task-bootstrap confidence intervals (4,000 resamples).

\begin{table*}[!htbp]
\centering
\caption{\textbf{Main-study evaluation groups.} Candidates include final gate failures (zeros in model/harness comparisons). Shared R1 baselines are counted once, yielding 1,010 candidates and 979 human-rated playable versions across 135 unique tasks.}
\label{tab:scale}\label{tab:app-candidates}
\apptab
\begin{tabularx}{\linewidth}{@{\hspace{5pt}}lrLrrr@{\hspace{5pt}}}
\apptop
\apphead {\bfseries\boldmath Evaluation group} & {\bfseries\boldmath Tasks} & {\bfseries\boldmath Executed comparison} & {\bfseries\boldmath Candidates} & {\bfseries\boldmath Human-rated} & {\bfseries\boldmath Gate failures} \\
\appmid
Easy & 30 & Eight models, OpenCode, R1 & 240 & 215 & 25 \\
\appzebra Medium & 40 & Four models, OpenCode, R1 & 160 & 159 & 1 \\
Hard & 50 & Two models, OpenCode, R1 & 100 & 98 & 2 \\
\appzebra Challenge R1 & 15 & Two models $\times$ five harnesses & 150 & 147 & 3 \\
Challenge R2--R4 & 15 & Astra $\times$ two harnesses $\times$ four strategies $\times$ three rounds & 360 & 360 & 0 \\
\appmid
\apphead {\bfseries\boldmath Total} & {\bfseries\boldmath 135 unique} & & {\bfseries\boldmath 1,010} & {\bfseries\boldmath 979} & {\bfseries\boldmath 31} \\
\appbot
\end{tabularx}
\end{table*}

\begin{table*}[!htbp]
\centering
\caption{\textbf{Single-round OpenCode results with 95\% task-bootstrap intervals.} Final gate failures enter as zeros.}
\label{tab:app-leaderboard-full}
\apptab
\begin{tabularx}{\linewidth}{@{\hspace{5pt}}lRRR@{\hspace{7pt}}lRRRRR@{\hspace{5pt}}}
\apptop
\apphead & \multicolumn{2}{c}{\bfseries Automated} & \multicolumn{3}{c}{\bfseries Human overall} & \multicolumn{4}{c}{\bfseries Human dimensions (0--25)} \\
\apphead {\bfseries\boldmath Model} & {\bfseries\boldmath Pass} & {\bfseries\boldmath Auto} & {\bfseries\boldmath RA} & {\bfseries\boldmath \appmuted{95\% CI}} & {\bfseries\boldmath $\mathrm{EC}$} & {\bfseries\boldmath Stab.} & {\bfseries\boldmath Play.} & {\bfseries\boldmath Visual} & {\bfseries\boldmath Inter.} \\
\appmid
\appgroupd{10}{Easy tier}{30 tasks, eight models} \\
\hspace{6pt}GPT-6 Astra & 100.0 & 81.95 & \appbest{46.61} & \appci{44.78, 48.61} & \appbest{36.43} & \appbest{10.57} & \appbest{8.57} & \appbest{8.43} & \appbest{8.85} \\
\hspace{6pt}Opus 5.5 & 100.0 & \appbest{86.75} & 40.94 & \appci{38.70, 43.02} & 30.32 & 8.61 & 8.06 & 6.54 & 7.11 \\
\hspace{6pt}Fable 5.1 & 100.0 & 73.62 & 37.46 & \appci{34.19, 40.46} & 28.00 & 8.33 & 6.87 & 5.79 & 7.01 \\
\hspace{6pt}Opus 5 & 76.7 & 52.98 & 29.17 & \appci{21.90, 36.48} & 22.54 & 6.13 & 5.64 & 5.60 & 5.17 \\
\hspace{6pt}Kimi K3 & 90.0 & 67.33 & 28.50 & \appci{23.63, 33.33} & 20.48 & 6.46 & 5.16 & 3.96 & 4.90 \\
\hspace{6pt}GPT-5.6 Sol & 96.7 & 71.87 & 26.85 & \appci{22.53, 31.19} & 20.77 & 7.10 & 4.83 & 3.96 & 4.88 \\
\hspace{6pt}GLM-5.3 & 83.3 & 60.43 & 24.45 & \appci{18.83, 29.64} & 17.67 & 5.39 & 4.25 & 3.67 & 4.35 \\
\hspace{6pt}DeepSeek V4-Flash & 70.0 & 50.98 & 13.66 & \appci{9.53, 17.93} & 10.03 & 3.53 & 2.28 & 1.98 & 2.25 \\
\appgroupd{10}{Medium tier}{40 tasks, four models} \\
\hspace{6pt}GPT-6 Astra & 100.0 & 77.16 & \appbest{48.21} & \appci{45.97, 50.47} & \appbest{39.02} & \appbest{11.37} & \appbest{9.14} & \appbest{8.86} & \appbest{9.65} \\
\hspace{6pt}Opus 5.5 & 97.5 & \appbest{80.63} & 38.99 & \appci{35.79, 41.76} & 29.25 & 8.40 & 7.61 & 6.32 & 6.93 \\
\hspace{6pt}Fable 5.1 & 100.0 & 76.89 & 37.64 & \appci{34.45, 40.87} & 28.64 & 8.63 & 7.26 & 5.92 & 6.82 \\
\hspace{6pt}Opus 5 & 100.0 & 75.45 & 35.20 & \appci{31.71, 38.48} & 25.87 & 7.08 & 6.44 & 6.10 & 6.25 \\
\appgroupd{10}{Hard tier}{50 tasks, two models} \\
\hspace{6pt}GPT-6 Astra & 100.0 & \appbest{87.61} & \appbest{42.71} & \appci{41.24, 44.23} & \appbest{33.24} & \appbest{9.40} & 7.91 & \appbest{7.98} & \appbest{7.95} \\
\hspace{6pt}Opus 5.5 & 96.0 & 83.68 & 40.37 & \appci{37.50, 42.87} & 31.08 & 8.89 & \appbest{8.11} & 6.69 & 7.40 \\
\appbot
\end{tabularx}
\end{table*}

\begin{table*}[!htbp]
\centering
\caption{\textbf{First-round results on 15 Challenge tasks.} DeepSeek denotes the DeepSeek CLI harness running each generator; final gate failures enter as zeros.}
\label{tab:app-harness-full}
\apptab\renewcommand{\arraystretch}{1.05}
\begin{tabularx}{\linewidth}{@{\hspace{5pt}}lRRR@{\hspace{7pt}}lRRRRR@{\hspace{5pt}}}
\apptop
\apphead & \multicolumn{2}{c}{\bfseries Automated} & \multicolumn{3}{c}{\bfseries Human overall} & \multicolumn{4}{c}{\bfseries Human dimensions (0--25)} \\
\apphead {\bfseries\boldmath Model or harness} & {\bfseries\boldmath Pass} & {\bfseries\boldmath Auto} & {\bfseries\boldmath RA} & {\bfseries\boldmath \appmuted{95\% CI}} & {\bfseries\boldmath $\mathrm{EC}$} & {\bfseries\boldmath Stab.} & {\bfseries\boldmath Play.} & {\bfseries\boldmath Visual} & {\bfseries\boldmath Inter.} \\
\appmid
\appgroup{10}{GPT-6 Astra} \\
\hspace{6pt}OpenCode & 100.0 & 87.95 & \appbest{41.85} & \appci{37.36, 46.06} & \appbest{32.54} & 9.77 & \appbest{7.91} & \appbest{6.97} & 7.89 \\
\hspace{6pt}Codex & 100.0 & 88.07 & 39.73 & \appci{34.30, 44.82} & 31.54 & \appbest{10.26} & 7.01 & 6.32 & \appbest{7.95} \\
\hspace{6pt}Pi & 93.3 & 81.88 & 36.80 & \appci{30.49, 41.83} & 28.23 & 8.32 & 6.75 & 6.40 & 6.76 \\
\hspace{6pt}DeepSeek & 100.0 & \appbest{88.72} & 35.70 & \appci{29.74, 40.82} & 28.12 & 8.52 & 6.58 & 6.13 & 6.89 \\
\hspace{6pt}Prime & 86.7 & 75.53 & 31.95 & \appci{23.29, 40.07} & 24.13 & 7.67 & 5.82 & 5.05 & 5.59 \\
\appgroup{10}{Opus 5.5} \\
\hspace{6pt}OpenCode & 100.0 & 89.33 & \appbest{46.71} & \appci{43.89, 49.57} & \appbest{36.79} & \appbest{10.47} & \appbest{9.71} & \appbest{7.17} & \appbest{9.43} \\
\hspace{6pt}Codex & 100.0 & \appbest{90.33} & 42.62 & \appci{38.91, 46.48} & 33.70 & 9.97 & 8.68 & 6.84 & 8.21 \\
\hspace{6pt}Pi & 100.0 & 86.63 & 40.96 & \appci{38.38, 43.36} & 31.37 & 9.24 & 8.08 & 6.35 & 7.70 \\
\hspace{6pt}Prime & 100.0 & 88.36 & 40.66 & \appci{37.56, 43.63} & 32.31 & 9.75 & 8.34 & 6.18 & 8.03 \\
\hspace{6pt}DeepSeek & 100.0 & 88.01 & 40.52 & \appci{37.78, 43.33} & 31.20 & 9.36 & 8.05 & 5.77 & 8.02 \\
\appgroup{10}{OpenCode follow-up} \\
\hspace{6pt}Fable 5.2 & 100.0 & 92.73 & 48.07 & [46.63, 49.49] & 37.65 & 9.83 & 9.60 & 8.80 & 9.41 \\
\hspace{6pt}GPT-6.2 Astra & 100.0 & 90.73 & 45.01 & [41.82, 47.89] & 34.26 & 8.84 & 8.06 & 9.08 & 8.28 \\
\hspace{6pt}Gemini 4 Neon & 93.3 & 86.92 & 38.18 & [31.71, 42.64] & 28.46 & 7.80 & 7.22 & 6.32 & 7.12 \\
\appbot
\end{tabularx}
\end{table*}

The follow-up added 45 OpenCode R1 builds on the same Challenge tasks with \texttt{max} reasoning effort for Fable~5.2 and GPT-6.2 Astra and \texttt{xhigh} for Gemini~4 Neon. Screening retained 640 of 648 ratings on the 44 playable builds. Across both studies, 11 generators produced 1,055 candidates on the same 135 tasks, with 1,023 playable versions and 12,434 qualified ratings.

For the reference-scope comparison, we compute Other$-$AAA within each model and evaluation tier, average models equally within tiers, then weight tiers by their task counts (30, 40, 50, 15). From Easy through Challenge, the within-tier RA gaps are 5.21, 5.21, 1.35, and 0.96 points. The 95\% percentile intervals use 10,000 task-bootstrap draws within each reference group. For the adjusted gaps, draws also preserve tier and model coverage, resampling each task jointly across models.

\textbf{Scope of the Challenge results.} Under OpenCode, Astra leads Opus~5.5 by 5.46 RA points across the 120 \pubedit{EEE} tasks. On the 15-task Challenge set, both models pass the runtime gates on all 15 under OpenCode, where Opus~5.5 wins 11 tasks and leads by 4.86 points, chiefly in Playability ($+1.80/25$) and Interactive Feedback ($+1.54/25$). The R4 player-feedback gains over \pubedit{Obj-RSI} ($8.92$ and $7.89$ RA points) apply to Astra under Codex and OpenCode, respectively.

\begin{table*}[!htbp]
\centering
\caption{\textbf{OpenCode Elo across the observed R1 schedule.} Each observed model pair has equal weight. Challenge ranks the five generators on the same 15 tasks. Each fit has mean Elo 1,500.}
\label{tab:opencode-ladder}
\apptab
\input{tab/opencode_ladder}
\end{table*}

{\normalcolor \textbf{Elo estimation.} The ladder (\cref{tab:opencode-ladder}) uses 6,611 of 6,834 ratings across 575 OpenCode R1 builds, with zero RA for 29 gate failures. On each shared task, 16,384 \href{https://doi.org/10.1214/aos/1176345338}{Bayesian bootstrap} draws reweight raters jointly across builds. Higher mean RA wins each draw, with half credit for ties. We average this credit over shared tasks for each of the 37 observed model pairs and fit an unpenalized \href{https://doi.org/10.1093/biomet/39.3-4.324}{Bradley--Terry} model. Elo is $1500+400\beta_m/\ln 10$ for centered log strength $\beta_m$. The 95\% intervals refit 4,000 within-tier task resamples with win credits fixed.}

\subsection{\pubedit{Multi-Round Feedback and Dimension Transitions}}
\label{sec:app-feedback-provider}\label{sec:app-feedback-results}\label{sec:app-experience-transitions}

\begin{figure*}[!htbp]
\centering
\includegraphics[width=0.92\linewidth]{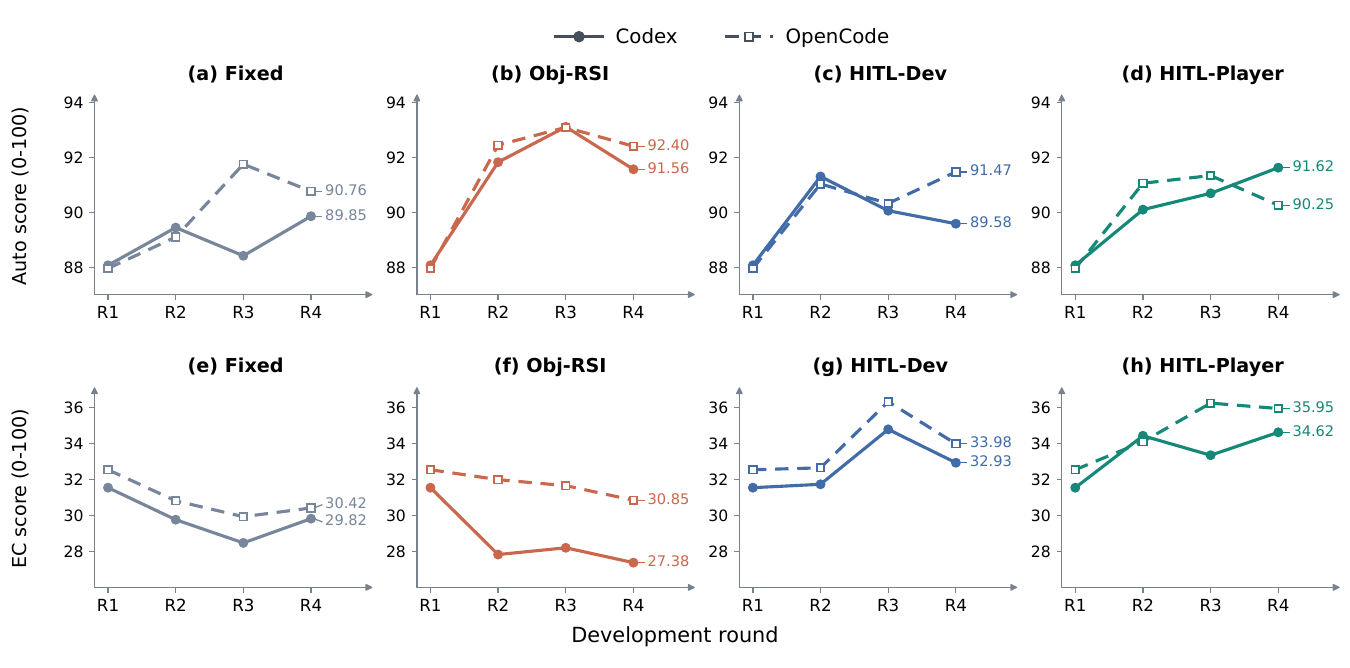}
\captionsetup{labelformat=pubappendix}
\caption{\textbf{Auto and Experience Composite ($\mathrm{EC}$) trajectories across R1--R4} on the 15 challenge tasks (solid: Codex; dashed: OpenCode).}
\label{fig:app-auto-ec-trajectories}
\end{figure*}

{\pubedit{\Cref{tab:app-multiround}}} reports R1/R4 means, raw and R1-SD-standardized gains, and adjacent-round tier regression rates across all eight strategy--harness branches; {\pubedit{\cref{fig:app-auto-ec-trajectories}}} plots the corresponding Auto and $\mathrm{EC}$ trajectories.
Of 420 HITL-Player R4 ratings, 159 came from players who had provided earlier feedback for the same task and harness. Excluding these providers from HITL-Player and \pubedit{Obj-RSI} retained all 15 tasks per harness.

In {\pubedit{Fig.~\ref{fig:experience-issues-revision}}} of the main paper, across the 390 versions in the multi-round Challenge experiment (30 shared R1 baselines and 360 R2--R4 revisions) and the 120 task--harness--dimension units at R1, 32 have median below B (12 Visual, 8 Playability, 8 Inter., 4 Stability) and 88 are B or above. At R4, HITL-Player recovers 29/32 sub-B units (10/12 Visual, 8/8 Playability, 7/8 Inter., 4/4 Stability) and retains 80/88 competent units; HITL-Dev ($n\ge 3$ developers) recovers 26/32 and retains 82/88; Fixed recovers 24/32 and retains 68/88; and \pubedit{Obj-RSI} recovers 7/32 while retaining only 60/88 (9/22 in Playability).

\begin{table*}[!htbp]
\centering
\captionsetup{labelformat=pubappendix}
\caption{\textbf{Multi-round feedback summary on 15 paired tasks.} R1 is shared across strategies within each harness. $\Delta$ is R4 minus R1; $\tilde\Delta_H$ and $\tilde\Delta_{\mathrm{Auto}}$ normalize gains by across-task R1 standard deviations within each harness. Regression rates cover 45 adjacent-round transitions per row.}
\label{tab:app-multiround}
\small
\setlength{\tabcolsep}{3pt}
\renewcommand{\arraystretch}{1.05}
\begin{tabular*}{\linewidth}{@{\extracolsep{\fill}}llrrcrrrr@{}}
\toprule
Harness & Strategy & R1 RA & R4 RA & $\Delta\mathrm{RA}$ & $\Delta\mathrm{Auto}$ & $\tilde\Delta_H$ & $\tilde\Delta_{\mathrm{Auto}}$ & Regress. (\%) \\
\midrule
Codex & Fixed & 39.73 & 39.85 & $+0.13$ & $+1.79$ & $+0.01$ & $+0.30$ & 40.0 \\
Codex & \pubedit{Obj-RSI} & 39.73 & 36.01 & $-3.72$ & $+3.49$ & $-0.35$ & $+0.59$ & 46.7 \\
Codex & HITL-Dev & 39.73 & 43.60 & $+3.88$ & $+1.51$ & $+0.36$ & $+0.25$ & 20.0 \\
Codex & HITL-Player & 39.73 & 44.92 & $+5.19$ & $+3.55$ & $+0.49$ & $+0.60$ & 31.1 \\
\midrule
OpenCode & Fixed & 41.85 & 41.23 & $-0.62$ & $+2.81$ & $-0.07$ & $+0.39$ & 31.1 \\
OpenCode & \pubedit{Obj-RSI} & 41.85 & 38.88 & $-2.97$ & $+4.45$ & $-0.34$ & $+0.61$ & 33.3 \\
OpenCode & HITL-Dev & 41.85 & 44.40 & $+2.55$ & $+3.51$ & $+0.29$ & $+0.48$ & 15.6 \\
OpenCode & HITL-Player & 41.85 & 46.77 & $+4.93$ & $+2.30$ & $+0.56$ & $+0.32$ & 20.0 \\
\bottomrule
\end{tabular*}
\end{table*}

\subsection{\pubedit{Rating Reliability and Auto Agreement}}
\label{sec:app-reliability}\label{sec:app-calibration}\label{sec:app-evaluation}

\pubedit{Removing 342 submissions retains} 11,794 qualified ratings \pubedit{and shifts} mean RA by only $0.06$ points ({\pubedit{\cref{tab:app-icc}}}). Unbalanced one-way ANOVA yields consensus reliability $\mathrm{ICC}(1,k_0)=0.727$ at $k_0=12.05$~\cite{shrout1979intraclass}. {\pubedit{\Cref{tab:app-agreement}}} confirms that Auto checklist totals correlate weakly with human RA across all rated subsets and turn slightly negative above $\mathrm{Auto}\ge 90$.

Across qualified sessions, active play time correlated weakly with the RA rating given in that session (Spearman $\rho=0.160$).

\begin{table}[!htbp]
\centering
\captionsetup{labelformat=pubappendix}
\caption{\textbf{Consensus reliability and screening summary} across the 979 playable versions (63 active annotators, $\ge 10$ ratings per version).}
\label{tab:app-icc}\label{tab:app-selection}
\apptab
\begin{tabularx}{\linewidth}{@{\hspace{5pt}}lRR@{\hspace{5pt}}}
\apptop
\apphead {\bfseries\boldmath Measure} & {\bfseries\boldmath Raw} & {\bfseries\boldmath Qualified} \\
\appmid
Retained ratings & 12,136 & 11,794 \\
\appzebra Equal-version mean raw $P$ & 39.92 & 39.86 \\
Harmonic replication $k_0$ & 12.40 & 12.05 \\
\appzebra Consensus $\mathrm{ICC}(1,k_0)$ & 0.711 & 0.727 \\
Active playtest time (person-h) & 332.6 & 321.0 \\
\appbot
\end{tabularx}
\end{table}

\begin{table}[!htbp]
\centering
\captionsetup{labelformat=pubappendix}
\caption{\textbf{Auto--human RA correlation} across subsets of main-study versions (gate-failure zeros excluded; 95\% task-cluster bootstrap CI).}
\label{tab:app-agreement}
\apptab
\begin{tabularx}{\linewidth}{@{\hspace{5pt}}lRRlR@{\hspace{5pt}}}
\apptop
\apphead {\bfseries\boldmath Scope} & {\bfseries\boldmath Versions} & {\bfseries\boldmath $\rho$} & {\bfseries\boldmath \appmuted{95\% CI}} & {\bfseries\boldmath $\tau_b$} \\
\appmid
All rated & 979 & 0.174 & \appci{0.087, 0.260} & 0.116 \\
\appzebra Single-shot & 472 & 0.192 & \appci{0.103, 0.278} & 0.123 \\
Challenge & 507 & 0.034 & \appci{$-0.055$, 0.134} & 0.024 \\
\appzebra $\mathrm{Auto}\geq80$ & 715 & 0.025 & \appci{$-0.070$, 0.116} & 0.017 \\
$\mathrm{Auto}\geq90$ & 429 & $-0.078$ & \appci{$-0.218$, 0.076} & $-0.055$ \\
\appzebra $\mathrm{Auto}\geq95$ & 46 & $-0.351$ & \appci{$-0.595$, 0.058} & $-0.247$ \\
\appbot
\end{tabularx}
\end{table}

\subsection{\pubedit{Reviewer Coverage and Leaderboard Sensitivity}}
\label{sec:app-sensitivity}

Autonomous play covers 118 DEV and 116 TEST versions, yielding 233 and 162 separated pairs, respectively.

{\pubedit{For Tables~\ref{tab:leaderboard}--\ref{tab:multiround}, applying}} $y_i^{\mathrm{GRA}}$ to the same qualified votes, with equal task weights and gate failures scored zero, keeps Astra first on Easy, Medium, and Hard. Its leads over Opus~5.5 are 3.02, 6.95, and 2.01 points (Holm-adjusted $p=0.137,\,0.0005,\,0.178$, respectively). At R4, HITL-Player still ranks first under both harnesses, ahead of \pubedit{Obj-RSI} by 7.33 points under Codex ($p_{\rm Holm}<0.001$) and 3.95 under OpenCode ($p_{\rm Holm}=0.131$).

\clearpage

\section{\pubedit{Supplementary Case Studies}}
\label{sec:app-case-studies}

\subsection{\pubedit{Generated Scenes and Player-Feedback Revisions}}
\label{sec:app-feedback-cases}

In the two illustrative R1$\to$R2 player-feedback cases ({\pubedit{Fig.~\ref{fig:experience-issues-revision}c,d}}), \emph{The Great Ace Attorney Chronicles} restores multi-round cross-examination and fact reconstruction, while \emph{Need for Speed Unbound} fixes inverted A/D steering and adds directional cues.

\begin{figure*}[!htbp]
\centering
\includegraphics[width=0.92\linewidth]{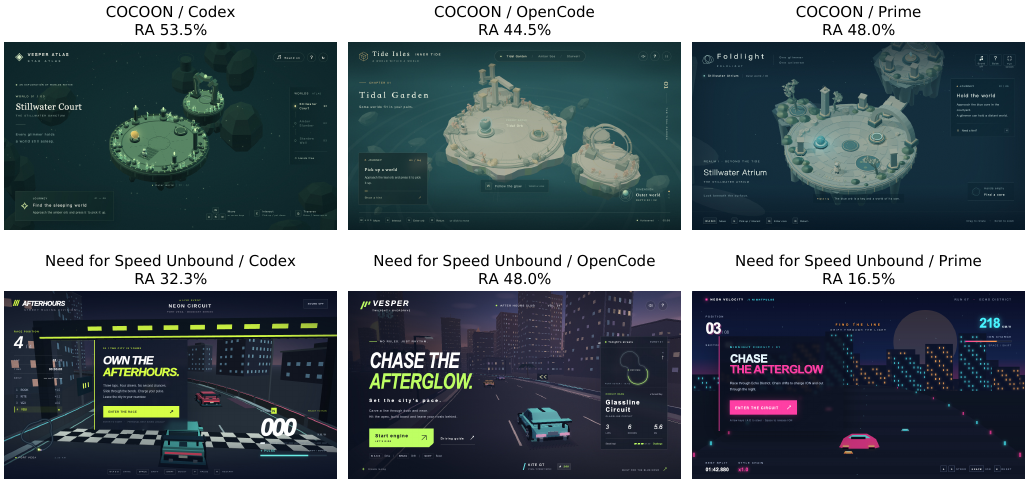}
\captionsetup{labelformat=pubappendix}
\caption{Illustrative R1 scenes for COCOON (top) and Need for Speed Unbound (bottom) generated by Astra under Codex, OpenCode, and Prime (left to right), labeled with qualified human RA means.}
\label{fig:qualitative}
\end{figure*}

\subsection{\pubedit{Qualitative Method Comparisons}}
\label{sec:app-qualitative-methods}

\begin{figure*}[!htbp]
  \centering
  \captionsetup{position=top}
  \caption{\textbf{Two generators at the same opening countdown in \emph{Gang Beasts}.}
  Both Hard R1 builds use OpenCode. At the count of two, GPT-6 Astra shows the entire rooftop inside a control interface, while Opus~5.5 places the camera close to the fighters. Playtesters rated Astra higher, but Auto scored Opus~5.5 higher.}
  \label{fig:app-case-rooftop}
  \includegraphics[width=\linewidth]{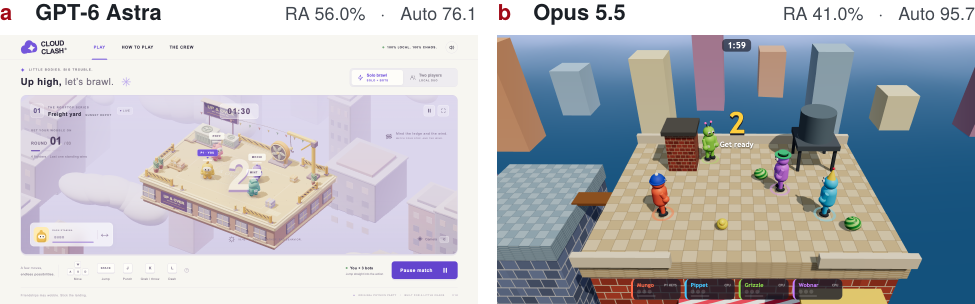}
\end{figure*}

\begin{figure*}[!htbp]
  \centering
  \captionsetup{position=top}
  \caption{\textbf{Two starting parks in \emph{Planet Zoo}.}
  Given the same Hard R1 prompt under OpenCode, GPT-6 Astra opens on populated enclosures with a management panel. Opus~5.5 starts the player on an open plot with a path and construction tools. Playtesters gave the latter version the higher RA.}
  \label{fig:app-case-zoo}
  \includegraphics[width=\linewidth]{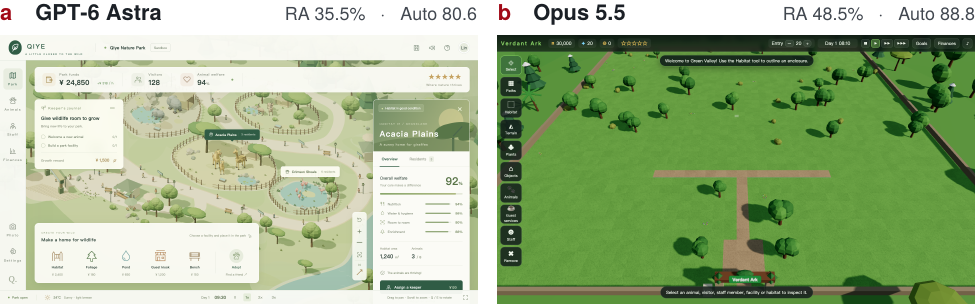}
\end{figure*}

\begin{figure*}[!htbp]
  \centering
  \captionsetup{position=top}
  \caption{\textbf{Two harnesses render the same strategy prompt.}
  For the \emph{Total War: THREE KINGDOMS} challenge, GPT-6 Astra with Codex places settlements and faction banners on a continuous hex map. With Prime, labeled territories sit on separate tiles. Both R1 builds pass the functional gates.}
  \label{fig:app-case-harness}
  \includegraphics[width=\linewidth]{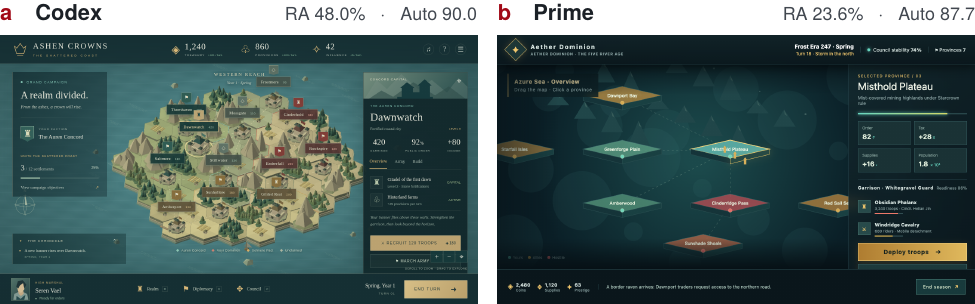}
\end{figure*}

\begin{figure*}[!htbp]
  \centering
  \captionsetup{position=top}
  \caption{\textbf{Two feedback revisions of the opening course in \emph{Thumper}.}
  The shared Astra/Codex R1 build scored 31.8\% RA. After developer feedback, a four-step tutorial waits for a timed keypress at the hit line (47.3\% RA). After player feedback, a timing gauge and a bright impact ring make successful hits easier to see (41.4\% RA).}
  \label{fig:app-case-rhythm}
  \includegraphics[width=\linewidth]{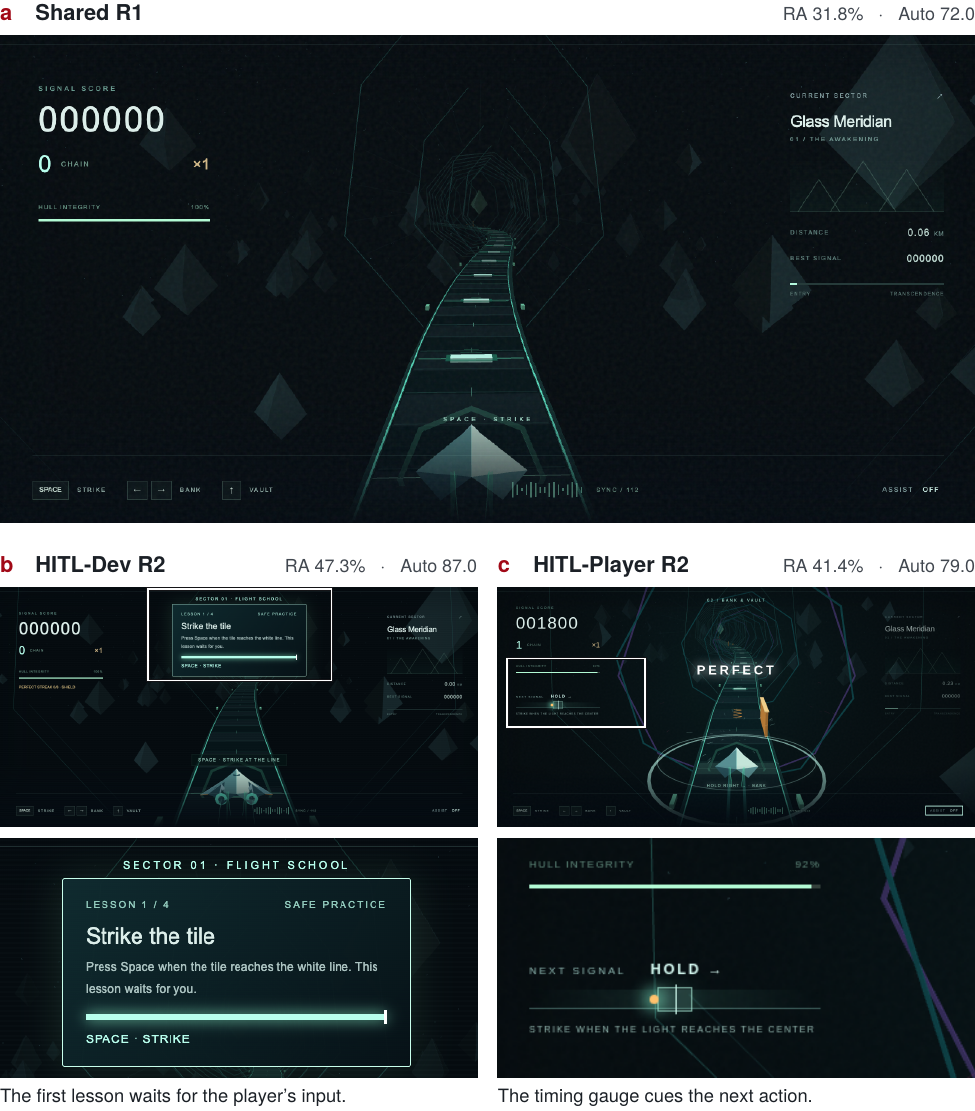}
\end{figure*}

\begin{figure*}[!htbp]
  \centering
  \captionsetup{position=top}
  \caption{\textbf{What the player can see before attacking in \emph{Persona~5 Royal}.}
  Four R4 feedback strategies branch from one Astra/Codex build. Each frame shows the first combat turn. The boxed interface details are enlarged below each frame. \pubedit{Obj-RSI} has the highest Auto score here, while HITL-Player has the highest RA.}
  \label{fig:app-case-battle}
  \includegraphics[width=\linewidth]{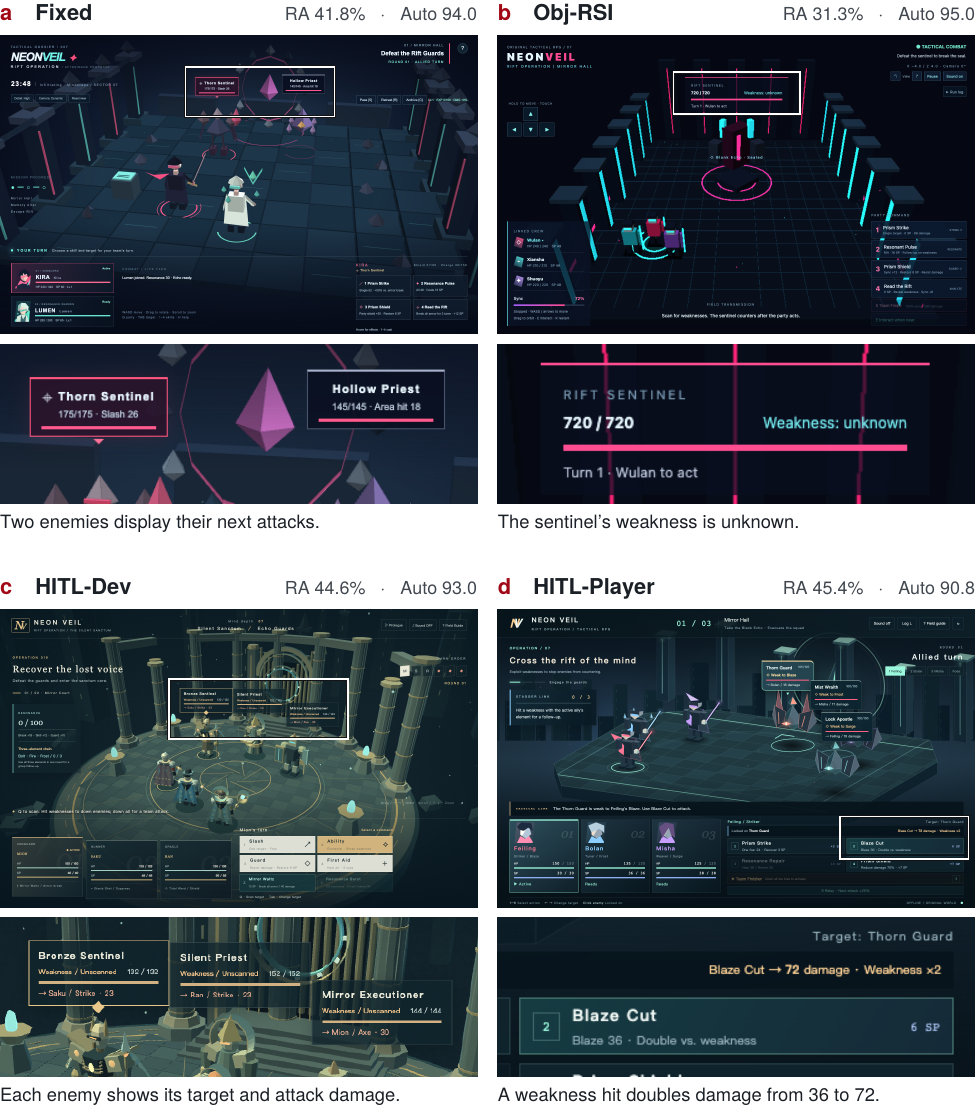}
\end{figure*}

\Cref{fig:app-case-rooftop,fig:app-case-zoo,fig:app-case-harness} compare generators and harnesses on matched prompts. In \emph{Thumper}, both feedback branches improve on the same R1 build (\cref{fig:app-case-rhythm}). \Cref{fig:app-case-battle} compares combat cues across four R4 branches.

\FloatBarrier

%% file: tab/opencode_ladder.tex
\begin{tabularx}{\linewidth}{@{\hspace{4pt}}r >{\raggedright\arraybackslash}X r r r >{\centering\arraybackslash}X r r@{\hspace{4pt}}}
\apptop
\apphead & & \multicolumn{2}{c}{Observed schedule} & \multicolumn{2}{c}{Equal model pairs} & \multicolumn{2}{c}{Challenge only} \\
\apphead \textbf{Rank} & \textbf{Generator} & \textbf{Tasks} & \textbf{Opponents} & \textbf{Elo} & \textbf{95\% CI} & \textbf{Rank} & \textbf{Elo} \\
\appmid
1 & Fable 5.2 & 15 & 4 & \textbf{1788.6} & [1719, 1876] & 1 & 1619.2 \\
\appzebra 2 & GPT-6.2 Astra & 15 & 4 & \textbf{1692.5} & [1588, 1812] & 3 & 1521.4 \\
3 & GPT-6 Astra & 135 & 10 & \textbf{1685.1} & [1648, 1732] & 4 & 1428.4 \\
\appzebra 4 & Opus 5.5 & 135 & 10 & \textbf{1657.1} & [1613, 1709] & 2 & 1570.7 \\
5 & Fable 5.1 & 70 & 7 & \textbf{1535.7} & [1483, 1592] & --- & --- \\
\appzebra 6 & Gemini 4 Neon & 15 & 4 & \textbf{1534.2} & [1419, 1640] & 5 & 1360.3 \\
7 & Opus 5 & 70 & 7 & \textbf{1414.5} & [1329, 1497] & --- & --- \\
\appzebra 8 & Kimi K3 & 30 & 7 & \textbf{1382.7} & [1300, 1453] & --- & --- \\
9 & GPT-5.6 Sol & 30 & 7 & \textbf{1339.9} & [1263, 1415] & --- & --- \\
\appzebra 10 & GLM-5.3 & 30 & 7 & \textbf{1331.3} & [1248, 1406] & --- & --- \\
11 & DeepSeek V4-Flash & 30 & 7 & \textbf{1138.4} & [1052, 1203] & --- & --- \\
\appbot
\end{tabularx}

%% file: main.bib
@inproceedings{bruce2024genie,
  author    = {Bruce, Jake and Dennis, Michael D. and Edwards, Ashley and Parker-Holder, Jack and Shi, Yuge and Hughes, Edward and others},
  title     = {{Genie}: Generative Interactive Environments},
  booktitle = {International Conference on Machine Learning (ICML)},
  series    = {Proceedings of Machine Learning Research},
  volume    = {235},
  year      = {2024}
}

@inproceedings{chen2024mllmjudge,
  author    = {Chen, Dongping and Chen, Ruoxi and Zhang, Shilin and others},
  title     = {{MLLM-as-a-Judge}: Assessing Multimodal {LLM}-as-a-Judge with Vision-Language Benchmark},
  booktitle = {International Conference on Machine Learning (ICML)},
  series    = {Proceedings of Machine Learning Research},
  volume    = {235},
  year      = {2024}
}

@article{chen2026gamexpert,
  author  = {Chen, Kun and Hong, Haorong and Gao, Peizhong and Lin, Jianfeng and Luo, Tongxu and others},
  title   = {{GameXpert-Bench}: How Far Are Coding Agents from Expert Game Development?},
  journal = {arXiv preprint arXiv:2608.21833},
  year    = {2026}
}

@article{chen2026swegame,
  author  = {Chen, Xiaoyu and Wei, Lai and Wang, Jin and Zou, Xiangyu and Fan, Ruochen and others},
  title   = {{SWE-Game}: Can Coding Agents Build the Games We Want?},
  journal = {arXiv preprint arXiv:2609.33678},
  year    = {2026}
}

@article{chi2026gamedevbench,
  author  = {Chi, Wayne and Fang, Yixiong and Yayavaram, Arnav and Yayavaram, Siddharth and Karten, Seth and Wei, Qiuhong Anna and Chen, Runkun and Wang, Alexander and Chen, Valerie and Talwalkar, Ameet and Donahue, Chris},
  title   = {{GameDevBench}: Evaluating Agentic Capabilities Through Game Development},
  journal = {arXiv preprint arXiv:2602.11103},
  year    = {2026}
}

@inproceedings{gao2023scaling,
  author    = {Gao, Leo and Schulman, John and Hilton, Jacob},
  title     = {Scaling Laws for Reward Model Overoptimization},
  booktitle = {International Conference on Machine Learning (ICML)},
  series    = {Proceedings of Machine Learning Research},
  volume    = {202},
  year      = {2023}
}

@inproceedings{he2024videoscore,
  author    = {He, Xuan and Jiang, Dongfu and Zhang, Ge and others},
  title     = {{VideoScore}: Building Automatic Metrics to Simulate Fine-grained Human Feedback for Video Generation},
  booktitle = {Conference on Empirical Methods in Natural Language Processing (EMNLP)},
  pages     = {2105--2123},
  year      = {2024}
}

@inproceedings{hu2024scenecraft,
  author    = {Hu, Ziniu and Iscen, Ahmet and Jain, Aashi and Kipf, Thomas and Yue, Yisong and Ross, David A. and Schmid, Cordelia and Fathi, Alireza},
  title     = {{SceneCraft}: An {LLM} Agent for Synthesizing {3D} Scenes as {Blender} Code},
  booktitle = {International Conference on Machine Learning (ICML)},
  series    = {Proceedings of Machine Learning Research},
  volume    = {235},
  year      = {2024}
}

@article{huang2026playtestarena,
  author  = {Huang, Yixu and Li, Bo and Li, Na and Wang, Zhe and Chen, Kaijie and others},
  title   = {{GUI} Agents for Continual Game Generation},
  journal = {arXiv preprint arXiv:2605.28258},
  year    = {2026}
}

@article{jia2026gamegenverifier,
  author  = {Jia, Chaobo and Wan, Ruipeng and Sun, Ting and Tan, Weihao and Wan, Borui and Tong, Yuxuan and Sheng, Guangming and Xu, Hong},
  title   = {{GameGen-Verifier}: Parallel Keypoint-Based Verification for {LLM}-Generated Games via Runtime State Injection},
  journal = {arXiv preprint arXiv:2605.07442},
  year    = {2026}
}

@article{jiang2026opengame,
  author  = {Jiang, Yilei and Hu, Jinyuan and Xiao, Qianyin and Zheng, Yaozhi and Ma, Ruize and others},
  title   = {{OpenGame}: Open Agentic Coding for Games},
  journal = {arXiv preprint arXiv:2604.18394},
  year    = {2026}
}

@article{kanervisto2025wham,
  author  = {Kanervisto, Anssi and Bignell, Dave and Wen, Linda Yilin and others},
  title   = {World and Human Action Models towards Gameplay Ideation},
  journal = {Nature},
  volume  = {638},
  number  = {8051},
  pages   = {656--663},
  year    = {2025}
}

@inproceedings{kim2020gamegan,
  author    = {Kim, Seung Wook and Zhou, Yuhao and Philion, Jonah and Torralba, Antonio and Fidler, Sanja},
  title     = {Learning to Simulate Dynamic Environments with {GameGAN}},
  booktitle = {IEEE/CVF Conference on Computer Vision and Pattern Recognition (CVPR)},
  pages     = {1228--1237},
  year      = {2020}
}

@article{lee2026a2z,
  author  = {Lee, Seonho and Jeong, Wonryeol and Cereser, Alberto and Kang, Inha and Kim, Hyeonjong and Kwak, Seungmin and Park, Dongmin},
  title   = {{A2Z GameSpec-Bench}: How Faithfully Can Coding Agents Generate Games from Game Design Specifications?},
  journal = {arXiv preprint arXiv:2609.39564},
  year    = {2026}
}

@inproceedings{li2025worldmodelbench,
  author    = {Li, Dacheng and Fang, Yunhao and Chen, Yukang and Yang, Shuo and Cao, Shiyi and Wong, Justin and Luo, Michael and Wang, Xiaolong and Yin, Hongxu and Gonzalez, Joseph E. and Stoica, Ion and Han, Song and Lu, Yao},
  title     = {{WorldModelBench}: Judging Video Generation Models As World Models},
  booktitle = {Advances in Neural Information Processing Systems (NeurIPS), Datasets and Benchmarks Track},
  year      = {2025}
}

@article{lu2026worldcoder,
  author  = {Lu, Shuo and Xu, Yinuo and Yu, Kecheng and Jiang, Siru and Yu, Yongcan and others},
  title   = {{WorldCoder-Bench}: Benchmarking Physically Grounded {3D} World Synthesis},
  journal = {arXiv preprint arXiv:2606.01869},
  year    = {2026}
}

@article{luo2026gamecraft,
  author  = {Luo, Tongxu and Wang, Rongsheng and Bi, Jiaxi and Xu, Chenming and Tang, Zhengyang and others},
  title   = {{GameCraft-Bench}: Can Agents Build Playable Games End-to-End in a Real Game Engine?},
  journal = {arXiv preprint arXiv:2606.17861},
  year    = {2026}
}

@inproceedings{panickssery2024llm,
  author    = {Panickssery, Arjun and Bowman, Samuel R. and Feng, Shi},
  title     = {{LLM} Evaluators Recognize and Favor Their Own Generations},
  booktitle = {Advances in Neural Information Processing Systems (NeurIPS)},
  year      = {2024}
}

@inproceedings{peng2026playcoder,
  author        = {Peng, Zhiyuan and Tao, Wei and Yin, Xin and Ying, Chenhao and Luo, Yuan and Guo, Yiwen},
  title         = {{PlayCoder}: Making {LLM}-Generated {GUI} Code Playable},
  booktitle     = {ACM International Conference on the Foundations of Software Engineering (FSE)},
  year          = {2026},
  eprint        = {2604.19742},
  archivePrefix = {arXiv}
}

@article{shrout1979intraclass,
  author  = {Shrout, Patrick E. and Fleiss, Joseph L.},
  title   = {Intraclass Correlations: Uses in Assessing Rater Reliability},
  journal = {Psychological Bulletin},
  volume  = {86},
  number  = {2},
  pages   = {420--428},
  year    = {1979}
}

@inproceedings{si2025design2code,
  author    = {Si, Chenglei and Zhang, Yanzhe and Li, Ryan and Yang, Zhengyuan and Liu, Ruibo and Yang, Diyi},
  title     = {{Design2Code}: Benchmarking Multimodal Code Generation for Automated Front-End Engineering},
  booktitle = {Conference of the Nations of the Americas Chapter of the Association for Computational Linguistics (NAACL)},
  pages     = {3956--3974},
  year      = {2025}
}

@inproceedings{valevski2025gamengen,
  author    = {Valevski, Dani and Leviathan, Yaniv and Arar, Moab and Fruchter, Shlomi},
  title     = {Diffusion Models Are Real-Time Game Engines},
  booktitle = {International Conference on Learning Representations (ICLR)},
  year      = {2025}
}

@inproceedings{xu2023imagereward,
  author    = {Xu, Jiazheng and Liu, Xiao and Wu, Yuchen and others},
  title     = {{ImageReward}: Learning and Evaluating Human Preferences for Text-to-Image Generation},
  booktitle = {Advances in Neural Information Processing Systems (NeurIPS)},
  year      = {2023}
}

@inproceedings{yang2024holodeck,
  author    = {Yang, Yue and Sun, Fan-Yun and Weihs, Luca and others},
  title     = {{Holodeck}: Language Guided Generation of {3D} Embodied {AI} Environments},
  booktitle = {IEEE/CVF Conference on Computer Vision and Pattern Recognition (CVPR)},
  pages     = {16277--16287},
  year      = {2024}
}

@article{zhang2025artifactsbench,
  author  = {Zhang, Chenchen and Li, Yuhang and Xu, Can and Liu, Jiaheng and Liu, Ao and others},
  title   = {{ArtifactsBench}: Bridging the Visual-Interactive Gap in {LLM} Code Generation Evaluation},
  journal = {arXiv preprint arXiv:2507.04952},
  year    = {2025}
}

@inproceedings{zhang2026vgamegym,
  author    = {Zhang, Wei and Yang, Jian and Tao, Renshuai and Chai, Linzheng and Guo, Shuyue and others},
  title     = {{V-GameGym}: Visual Game Generation for Code Large Language Models},
  booktitle = {Findings of the Association for Computational Linguistics: ACL 2026},
  pages     = {5613--5641},
  year      = {2026}
}

@article{zhang2026webgamebench,
  author  = {Zhang, Wenyu and You, Guoliang and Tianlun and Zhao, Haotian and Zhu, Tianshu and others},
  title   = {{WebGameBench}: Requirement-to-Application Evaluation for Coding Agents via Browser-Native Games},
  journal = {arXiv preprint arXiv:2605.17637},
  year    = {2026}
}

@article{zhang2026gameasg,
  author  = {Zhang, Xiuhui and Chen, Yi and Xu, Shusheng and Li, Fan and Wang, Huan and Yang, Tongkai and Yuan, Binhang},
  title   = {{GameASG-Bench}: Benchmarking Autonomous Software Generation for Game Development},
  journal = {arXiv preprint arXiv:2609.21293},
  year    = {2026}
}

@inproceedings{zhuge2025agentjudge,
  author    = {Zhuge, Mingchen and Zhao, Changsheng and Ashley, Dylan R. and others},
  title     = {{Agent-as-a-Judge}: Evaluate Agents with Agents},
  booktitle = {International Conference on Machine Learning (ICML)},
  series    = {Proceedings of Machine Learning Research},
  volume    = {267},
  year      = {2025}
}

@article{zhou2026agentic,
  title={Agentic Game Development as a Verifiable Trajectory Data Engine for Scaling World Models},
  author={Zhou, Pengfei and Wang, Hexin and Zhang, Zhengfeiyang and Ma, Yixing and Wan, Zhenglin and Zhang, Kaipeng and Zhao, Wangbo and You, Yang},
  journal={arXiv preprint arXiv:2608.25518},
  year={2026}
}

@article{luo2026ai,
  title={AI for games in the foundation model era},
  author={Luo, Meng and Li, Yanlin and Li, Hao and Lin, Hongzhan and Zhou, Pengfei and Ju, Tianjie and Zhang, Ran and Jin, Yeying and Lee, Mong-Li and Hsu, Wynne},
  journal={arXiv preprint arXiv:2609.16679},
  year={2026}
}
